\PassOptionsToPackage{table}{xcolor}
\documentclass[fleqn,10pt]{wlscirep}
\usepackage[utf8]{inputenc}
\usepackage[T1]{fontenc}
\usepackage{lineno}
\usepackage{xspace}
\usepackage{comment}
\usepackage{wrapfig}

\usepackage[export]{adjustbox}
\usepackage{booktabs}
\usepackage{graphicx}
\usepackage{subcaption}
\usepackage{enumitem}
\usepackage{fancyhdr}
\usepackage{textcomp}
\usepackage{expdlist}
\usepackage{gensymb}
\usepackage{layouts}
\usepackage{ccicons}
\usepackage{todonotes}
\usepackage{multirow}

\usepackage{tikz}
\usetikzlibrary{calc,tikzmark}   
\usepackage{tabu}                      
\usepackage{booktabs}                  
\usepackage{lipsum}                    
\usepackage{mwe}                       

\usepackage{xcolor}
\usepackage{graphicx}
\usepackage{subcaption}
\usepackage{float}

\usepackage{calc}

\newcommand{\dataname}{PathoGaze1.0\xspace}

\newcommand{\cczerolink}
{\href{https://creativecommons.org/public-domain}{the Creative Commons Public Domain Dedication license \mbox{\ccLogo\,\ccZero}}}

\newcommand{\ccfourlink}
{\href{https://creativecommons.org/licenses/by/4.0/}{Creative Commons Attribution 4.0 International License \mbox{\ccLogo\,\ccAttribution}}}

\newcommand{\eg}{e.g.\xspace}
\newcommand{\ie}{i.e.\xspace}
\newcommand{\etc}{etc.}

\newcommand{\eonedataname}{{P10S60T600}\xspace}

\newcommand{\etwodataname}{P9D397T540\xspace}

\newcommand{\totalreadingcount}{1,140\xspace}
\newcommand{\totaleonetrialcount}{600\xspace}
\newcommand{\totaletwotrialcount}{540\xspace}

\newcommand{\totalwsicount}{397\xspace}
\newcommand{\totaleonewsicount}{60\xspace}
\newcommand{\totaletwowsicount}{397\xspace}

\newcommand{\totaluniquepathologistparticipations}{11\xspace}
\newcommand{\totaleonepathologists}{10\xspace}
\newcommand{\totaletwopathologists}{9\xspace}

\newcommand{\totalhourcount}{18.69\xspace} 
\newcommand{\totaleonehourcount}{9.55\xspace} 
\newcommand{\totaletwohourcount}{9.14\xspace} 

\newcommand{\totalfixations}{171,909\xspace} 
\newcommand{\totaleonefixations}{86,304\xspace} 
\newcommand{\totaletwofixations}{85,605\xspace}

\newcommand{\totaletwoviewports}{254,369\xspace} 

\definecolor{seccolor}{RGB}{0, 128, 128}
\definecolor{subseccolor}{RGB}{10, 10, 128}
\definecolor{jccolor}{RGB}{128, 128, 28}

\definecolor{tablebold}{RGB}{150,0,0}

\newcommand{\scolor}[1]{\textcolor{seccolor}{#1}}

\newcommand{\sscolor}[1]{\textcolor{subseccolor}{#1}}

\newcommand{\testbedname}{PTAH\xspace}

\definecolor{rv1color}{RGB}{0, 0, 0}
\newcommand{\rv}[1]{\textcolor{rv1color}{#1}}
\definecolor{rvvtcolor}{RGB}{0, 150, 0}
\newcommand{\rvvt}[1]{\textcolor{rv1color}{#1}}

\definecolor{rvjccolor}{RGB}{0, 0, 0}
\newcommand{\rvjc}[1]{\textcolor{rvjccolor}{#1}}

\newlength\myheight
\newlength\mydepth

\AtBeginDocument{
  \settoheight{\myheight}{Xygp}
  \settodepth{\mydepth}{Xygp}
  \addtolength{\myheight}{\mydepth}
}
\newcommand*\wordimg[1]{
  \raisebox{-0.6\mydepth}{\includegraphics[height=0.6\myheight]{#1}}
}

\title{\dataname: An AI-Ready Navigation, Eye-Tracking, and Decision Dataset in Pathology 
}

\author[1,$\dag$]{Veronica Thai}
\author[2,$\dag$]{Rui Li}
\author[3,$\dag$]{Meng Ling}
\author[4]{Shuning Jiang}
\author[5]{Jeremy Wolfe}
\author[1]{Raghu Machiraju}
\author[6]{Yan Hu}
\author[6]{Zaibo Li}

\author[6,*]{Anil Parwani}
\author[1,*]{Jian Chen}
\affil[1]{Department of Computer Science and Engineering, The Ohio State University, Columbus, OH 43210, USA}
\affil[2]{MathWorks, Natick, MA 01760, USA}
\affil[3]{The Washington Post, Washington, DC 20071, USA}
\affil[4]{Google LLC, Mountain View, CA 94043, USA}
\affil[5]{Departments of Ophthalmology and Radiology, Harvard Medical School, and Department of Surgery, Brigham and Women's Hospital, Boston, MA 02115, USA}
\affil[6]{Department of Pathology, The Ohio State University Wexner Medical Center, Columbus, OH 43210, USA}

\affil[*]{Corresponding authors: Anil Parwani  (anil.parwani@osumc.edu) and Jian Chen (chen.8028@osu.edu)}

\affil[$\dag$]{These authors are co-first authors and contributed equally to this work.}

\begin{abstract}

Performing cancer diagnosis on giga-pixel whole-slide images (WSIs) is an important but difficult visual search task. \rv{The large giga-pixel image size forces pathologists to navigate by 
zooming and panning to search %and examine 
tissue for malignancy.}
Diagnostic accuracy
averages around $70\%$, and adding a second pathologist does not substantially improve consistencies between diagnoses. 
\rv{There is a lack of 
publicly available 
behavioral data 
to explain these diagnostic errors and inconsistencies. 
To fill this gap, we
curated 
\dataname,
a comprehensive 
behavioral dataset capturing 
high-fidelity clinical diagnostic workflows across \totaluniquepathologistparticipations~\rvvt{different} pathologist participants over two experiments 12-months apart (8 overlapping pathologists \rvvt{across both experiments}),
as they interpreted \totalwsicount~WSIs
from CAMELYON16, a set of publicly available breast cancer lymph node WSIs. 
\rvvt{The dataset} comprises \totalhourcount~hours of eye-tracking
(\totalfixations~fixations), %\totalsaccades~saccades) 
and navigational scanning data 
 along with \totalreadingcount~diagnostic decisions, all collected using a newly-developed testbed \testbedname. 
The testbed supports interaction techniques found in clinical tools and collects
viewport, mouse actions, and diagnostic decisions, enabling cross-links between WSI image stimulus, gaze, navigation behaviors, and diagnostic decisions. 
We share preregistered experimental materials, the datasets, and processing methods on the Open Science Framework (OSF).}
\end{abstract}

\begin{document}

\flushbottom
\maketitle

\section*{\scolor{Background \& Summary}}
\rv{The third-leading cause of death in the US is medical errors.
Among all diagnostic errors, $60-70\%$ are believed to be due to perceptual (detection) and cognitive (interpretive) errors~\cite{balogh2015improving, west2018physician}.
In pathology, 
%one of these errors 
%is
\textit{navigation error} can also occur, \ie, failing to bring a tumor region into the visible viewport
or failing to display the tissue at a resolution sufficient to see it~\cite{drew2021more}.
This occurs because pathologists’ visual search is dynamic~\cite{molin2015slide}, and in giga-pixel whole-slide images (WSIs), which are typically scanned at 20$\times$ magnification and 0.5~$\mu$m/pixel~\cite{parwani2021whole, forlines2009improving}, only a small proportion of the image area can fit on the screen. Pathologists must pan and zoom, \ie, navigating in
2.5D~\cite{plumlee2006zooming}, in image plane with depth-based zooming, while panning extensively at the same depth plane, to bring the correct portions
of the image into view at a useful
scale~\cite{plumlee2006zooming}.
%magnification. 
Such a diagnostic process is error-prone and the lack of real-world diagnostic consistency underscore the urgent need to capture %navigation
visual search behaviors and encode pathologists’ knowledge into reusable and computable forms for %downstream 
computational analyses.} 
%This motivates us to capture \textit{where}  pathologists look, \textit{how} their search differs, and reason \textit{why} and \textit{when} inconsistencies and errors occur on a large set of WSIs.

%\subsection*{\scolor{Related Work}}

\begin{table}[!t]
    \centering
    \scriptsize
    \setlength{\tabcolsep}{3pt}
    \resizebox{\textwidth}{!}{
    \begin{tabular}{p{0.5in} c c c c c c p{0.4in} p{0.55in} p{0.55in} p{0.9in} p{0.45in} p{0.45in} p{0.35in} p{0.3in} p{0.4in} p{0.45in} p{0.6in}}
        \toprule
        \multirow{2}{0.7in}{\textbf{Paper}} & \multicolumn{2}{c}{\textbf{Navigation/Action}} & \multicolumn{4}{c}{\textbf{Viewing}} & \multirow{2}{0.4in}{\textbf{Diagnosis}} & \multirow{2}{0.55in}{\textbf{Experience Level Label}} & \multirow{2}{0.55in}{\textbf{Instruction}} & \multirow{2}{0.9in}{\textbf{Purpose}} & \multirow{2}{0.45in}{\textbf{Eye Tracker}} & \multirow{2}{0.45in}{\textbf{Setting}} & \multirow{2}{0.35in}{\textbf{Tissue}} & \multirow{2}{0.3in}{\textbf{Slides}} & \multirow{2}{0.4in}{\textbf{\# of Experiments}} & \multirow{2}{0.45in}{\textbf{\# of Readings}} & \multirow{2}{0.55in}{\textbf{Data Availability}}\\
        \cmidrule(lr){2-3} \cmidrule(lr){4-7}
         & \textbf{Viewport} & \textbf{Mouse} & \textbf{Fixations} & \textbf{Saccades} & \textbf{Pupil} & \textbf{Frequency} & & & & & & & & & & \\
        \midrule
        Kockwelp et al. (2026)~\cite{kockwelp199martha} & -- & -- & $\checkmark$ & -- & -- & 120Hz & -- & -- & Draw segmentation mask & Use gaze data to train a model to facilitate slide annotation via gaze & Tobii Pro Fusion & Controlled, with instruction to draw mask & Testicular & 74 & 1 & --  & Available upon qualified request\\
        Lopes et al. (2026)~\cite{lopes2026exploring} & -- & -- & $\checkmark$ & -- & -- & 120Hz & -- & Pathologist & Verbally classify fixation cells during viewing & Use gaze data to train a model to facilitate slide annotation via gaze & Tobii Pro Fusion & Controlled, with instruction to classify fixations & Lung & 1 & 1 & 1 & Not stated\\
        Sprinkle et al. (2024)~\cite{sprinkle2024comparing} & -- & -- & $\checkmark$ & -- & -- & -- & Benign, atypical, malignant, defer & Resident, pathologist, cytologist & -- & Compare viewing patterns of residents and experts to understand how slides are examined & EyeLink 1000 & Controlled, with chinrest and forehead bar & Bladder & 20 & 1 & 80 & Not stated\\
        Nan et al. (2025)~\cite{nan2025deep} & $\checkmark$ & -- & $\checkmark$ & -- & -- & 60Hz & Benign (nevus), BCC, melanoma, SCC, SK & Pathologist & As natural as possible & Collect gaze patterns to train deep learning models & Tobii Pro Spectrum & Naturalistic & Skin & 1,565 & 1 & 3,978 & 150 data pairs publicly available, remainder upon formal request\\
        Drew et al. (2023)~\cite{drew2023pathologist} & $\checkmark$ & -- & $\checkmark$ & -- & $\checkmark$ & 250Hz & 5 diagnostic categories & Junior residents, senior residents, faculty & As natural as possible & Explore relation between pupil diameter and experience level and perceived difficulty of a slide & SMI RED-m & Controlled, with head position feedback & Breast & 32 & 1 & 1,138 & Upon request on a case-by-case basis\\
        Bruny\'{e} et al. (2023)~\cite{brunye2023image} & $\checkmark$ & -- & $\checkmark$ & $\checkmark$ & -- & 250Hz & 5 diagnostic categories & Residents, fellows, attendings & As natural as possible & Explore sources of diagnostic error & SMI RED250 & Naturalistic & Breast & 32 & 1 & 1,075 & Upon request on a case-by-case basis \\
        Bruny\'{e} et al. (2023)~\cite{brunye2023zoom} & $\checkmark$ & -- & $\checkmark$ & -- & $\checkmark$ & 250Hz & 5 diagnostic categories & Residents by year, fellows, practicing pathologists by $\leq$ 4, 5-9, 10-19, 20+ years & As natural as possible & Explore whether pupil size is associated with perceived difficulty & SMI RED & Naturalistic & Breast & 32 & 1 & 1,246 & Stated to be available, link broken at time of writing\\
        Mariam et al. (2022)~\cite{mariam2022smart} & -- & -- & $\checkmark$ & -- & -- & 60Hz & -- & Expert pathologists & Focus on lesions & Use gaze in place of hand drawn manual annotation for training ML models & Gazepoint GP3 & Controlled with instruction to focus on lesions & Oral & 4 & 1 & 8 & Publicly available\\
        Drew et al. (2021)~\cite{drew2021more} & $\checkmark$ & -- & $\checkmark$ & $\checkmark$ & -- & 250Hz & 5 diagnostic categories & Residents, fellows, attendings & As natural as possible & Explore the relationship between search behaviors and accuracy & SMI RED-m & Controlled, with head position feedback & Breast & 32 & 1 & 1,036 & Not stated\\
        Sudin et al. (2021)~\cite{sudin2021eye} & $\checkmark$ & -- & $\checkmark$ & -- & -- & 60Hz & 6 diagnostic categories & Certified general pathologists & As natural as possible & Explore relationship between reading duration and zoom patterns with experience & Smart Eye & Naturalistic & Breast & 20 & 1 & 60 & Not stated\\
        Kimeswenger et al. (2021)~\cite{kimeswenger2021artificial} & -- & -- & $\checkmark$ & -- & -- & 60/120Hz & 6 diagnostic categories & A qualified consultant and trainees & As natural as possible & Compare the regions of interest of pathologists with those of neural networks & iViewX RED & Naturalistic & Skin & 9 & 1 & 36 & Not stated\\
        Bruny\'{e} et al. (2017)~\cite{brunye2017accuracy} & $\checkmark$ & -- & $\checkmark$ & -- & -- & 60Hz & 5 diagnostic categories & Residents, faculty & As natural as possible & Explore relation between fixation features (duration, position relative to ROIs) and diagnostic accuracy and efficiency & SMI RED-m & Naturalistic & Breast & 24 & 1 & 480 & Not stated\\
        \midrule
        This paper & $\checkmark$ & $\checkmark$ & $\checkmark$ & $\checkmark$ & $\checkmark$ &  120 Hz and 60 Hz &  Benign, atypical, malignant, and neoplastic & Resident, $<$5, 5--10, $>$10 yrs & As natural as possible & Collect natural gaze data to support investigation of pathologists' behaviors & Tobii Pro Fusion & Naturalistic & Lymph node & 397 & 2 & 1,140 (600+540) & Publicly available\\
        \bottomrule
    \end{tabular}
    }
    \caption{\rv{\textbf{Viewport, Navigation, and Gaze Datasets in Digital Pathology.} 
    %Datasets remain limited in the total number of slides with ground-truth annotated regions for error analyses and AI-ready structured data. These small datasets highlight the challenge of collecting multimodal data for behavior studies in a high-fidelity clinical setting, with noise removed and data synchronized across multiple coordinates and tracking devices. This data collection fills this gap by capturing naturalistic behavior through a custom testbed and making these behaviors computable.
    A number of papers published earlier, developed independently, confirmed that gaze behavior, mouse behavior, and viewport behavior tracking are useful for downstream analyses and AI supervision. Reviewing these works can provide a more comprehensive understanding of multimodality behavior analyses. \textbf{Observation.} Our work scales up the data captures using a high-fidelity clinic settings in two experiments, with 12-month apart for an addition to the state-of-the-art pathologists' behavior studies.
    %especially consistency measure.  
    }}
    \label{tab:datasets}
\end{table}

%Beyond the behavior capturing, because of inconsistency and the gigapixel size of images from diverse diagnostic scenarios, AI-ready dataset construction is especially time-consuming and difficult to scale. 
\rv{While observing pathologists’ 
%naturalistic 
diagnostic process, 
%behaviors in the first experiment, 
we noticed that pathologists spent a seemingly limited amount of time %, around one minute, 
before making final diagnostic decisions. On the one hand, this search behavior is somewhat counterintuitive given the tissue complexity; on the other hand, it could mean that, as trained experts, pathologists’ fixations carry useful signals about tumorous tissues and their reasoning. 
A number of papers published earlier, all developed independently, confirmed both  observations. 
% Participants in Bejnordi et al~\cite{bejnordi2017diagnostic} had about the same task completion time. 
Mariam et al.~\cite{mariam2022smart} validated the use of gaze to supervise artificial intelligence (AI) in controlled lab settings. Chakraborty et al. showed that the viewport data provide sufficient supervision to train models for tumor detection~\cite{Chakraborty}. 
%where behaviors were constrained. 
Sprinkle et al.~\cite{sprinkle2024comparing}, Drew et al.~\cite{drew2023pathologist}, and Bruny\`{e} et al.~\cite{brunye2023image} %focuses on or repeated viewings on a smaller set of slides, which 
%allows one to examine for 
showed that errors are independent of 
%examined 
%behavior differences, 
zooming, 
%such as 
pupil diameter, and experience level. Lopes et al. provided extensive reviews of gaze variables and behavior analysis challenges~\cite{lopes2024eye}.
 Gaze capture beyond a two-dimensional (2D) screen tends to be less accurate than 2D capture with a chin-rest~\cite{drew2021more}. When the image cannot fit into the screen, scanpaths, or the sequence of eye movements, need to be combined with panning, to examine what tissue areas viewers see, because participants can fixate on the same screen location while using the keyboard or mouse to pan across the image space. 
 %Our most accurate participants adopted this behavior (Experiment I, Participant ID 4 [EI.p4]; Experiment II, Participant ID 3 [EII.p3]).
The reader is urged to review these works for a more complete understanding of the field of analyzing pathologists' complex visual search.}

\rv{These works and our observations motivated us to capture \textit{where} pathologists look, \textit{how} their search differs, and reason \textit{why} and \textit{when} inconsistencies and errors occur on a large set of WSIs.
This paper expands the data collection across a broader set of slides and instructed pathologists to search as they would in a clinical setting.
The validation of the use of 
%natural 
high-fidelity clinical behavior can  broaden the purview toward a future annotation-free machine learning supervision method that captures large-scale gaze behaviors in everyday clinical use, as well as reasoning why pathologists do not fixate on some tumorous tissue regions. 
%to supervise AI. 
%Motivated by this simple idea, 
%also leads to the second data capture experiment, which
We collected %uses an even 
a relatively large number of gaze behaviors for WSIs and prepared the gaze data to be AI-ready by following Croissant~\cite{akhtar2024croissant}, an open, standardized metadata format to make the datasets discoverable, portable, and interoperable; we also managed  noise, modality alignment, and derived behavior metric data so that they are useful for supervising AIs. 
%Other remain relatively small or
Compared to many state-of-the-art data collections in Table~\ref{tab:datasets}, our data were captured at scale, and more importantly, the acquisition method used
%can be extended to 
everyday clinical viewing settings, allowing a much larger scale data acquisition and comparative studies of behavioral responses to WSI stimuli.}
%between experiments.
%as about the same set of pathologists volunteered both experiments. 
%Most of the other datasets are complementary to this data, as they capture a few special-purpose behaviors.

%\rv{Others, such as but their experiment required participants to fixate on regions of interest for extended periods, which limits its ability for routine clinical uses.}

\begin{figure}[!t]
    \centering
%     \begin{minipage}[t]{0.7\linewidth}          % [t] instead of [c]: tops align
%         \makebox[\linewidth][l]{\textbf{a}}\\[0.2em]
%         \begin{tikzpicture}[remember picture]
%             \node[inner sep=0pt, anchor=north west] (imgA) {%
%                 \includegraphics[trim={240 50 460 50}, clip, width=\linewidth]{Figures/displayCoord.pdf}};
%         \end{tikzpicture}
%     \end{minipage}%
%     \hfill
%     \begin{minipage}[t]{0.28\linewidth}         % [t] here too
%         \makebox[\linewidth][l]{\textbf{b}}\\[0.2em]
%         \begin{tikzpicture}[remember picture]
%             \node[inner sep=0pt, anchor=north west] (imgB) {%
%                 \includegraphics[width=\linewidth]{Figures/interfaceExample_crop.png}};
%         \end{tikzpicture}

%         \vspace{0.8em}

%         \makebox[\linewidth][l]{\textbf{c}}\\[0.2em]
%         \includegraphics[width=\linewidth]{Figures/eyeTrackingResults/success_error2.pdf}
%     \end{minipage}

%    \begin{tikzpicture}[remember picture, overlay]
%     \draw[gray!60, line width=0.6pt]
%         \imgpt{imgB}{0.48}{0.151} -- \imgpt{imgA}{0.96}{0.1115};   % top: b screen -> a screen top-right
%     \draw[gray!60, line width=0.6pt]
%         \imgpt{imgB}{0.478}{0.45} -- \imgpt{imgA}{0.965}{0.725};   % bottom: b screen -> a screen bottom-right
% \end{tikzpicture}
    \includegraphics[width=\linewidth]{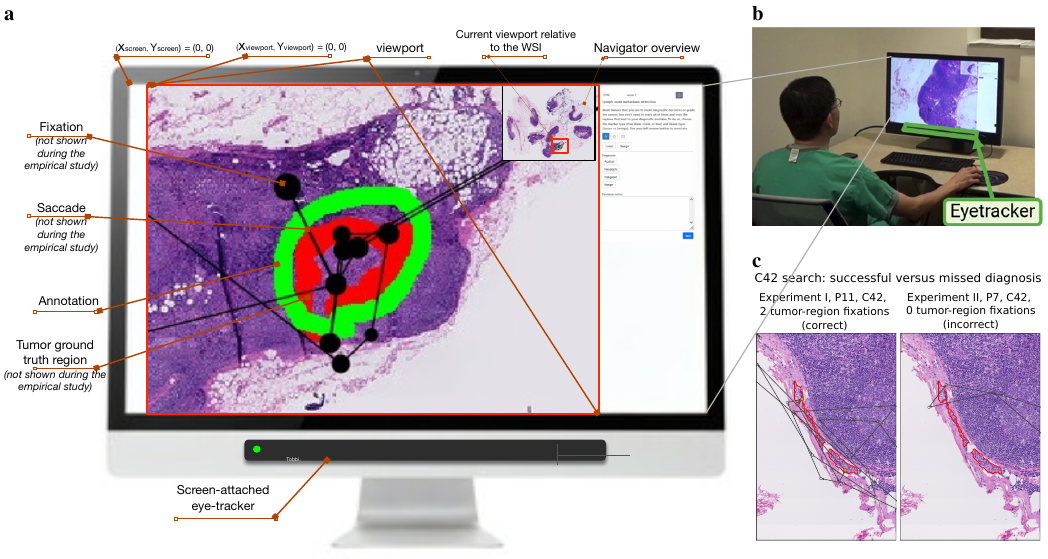}

 \caption{\rv{The \textbf{\testbedname testbed interface, coordinates, and workflow data collected for whole-slide image (WSI) diagnosis.} (a) The interface includes a main viewing panel, an overview thumbnail,
 %a resizable selection box indicating the current region of interest (ROI), 
 controls for panning and zooming, and an annotated region. 
    %The left panel shows the coordinate system. 
    The right pane illustrates the data-collection interface. Tumor regions are outlined in red\wordimg{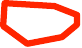}; The green curve illustrates an annotation made by a participant in the experiment; Black lines\wordimg{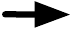} indicate saccade paths, circles indicate fixation locations.
    %and orange dots indicate fixations within tumor regions.
    The WSI coordinate origin is at the upper-left corner, $(0,0)$, and the bottom-right corner is $(x_{image}, y_{image})$, corresponding to the full pixel resolution shown in the thumbnail. (b) A photo of the data-collection environment. 
    %Because WSIs contain regions that vary in diagnostic appearance, patient characteristics, and hematoxylin and eosin ($H\&E$) staining quality, pathologists continually adjust the viewport and magnification during search. 
    The testbed records participants'  %interactions
    %together with
    eye gaze, mouse interaction, viewport, and diagnostic decisions, enabling cross-links between WSI image stimulus, gaze, navigation behaviors, and diagnostic decisions. 
    %comparisons across pathologists and repeated observations. 
    (c) The same participant examined the same slides in two separate experiments and showed different gaze behaviors.}}
    \label{fig:studySetupCoords}
\end{figure}

\begin{figure}[!t]
    \centering

    \includegraphics[width=0.99\linewidth]{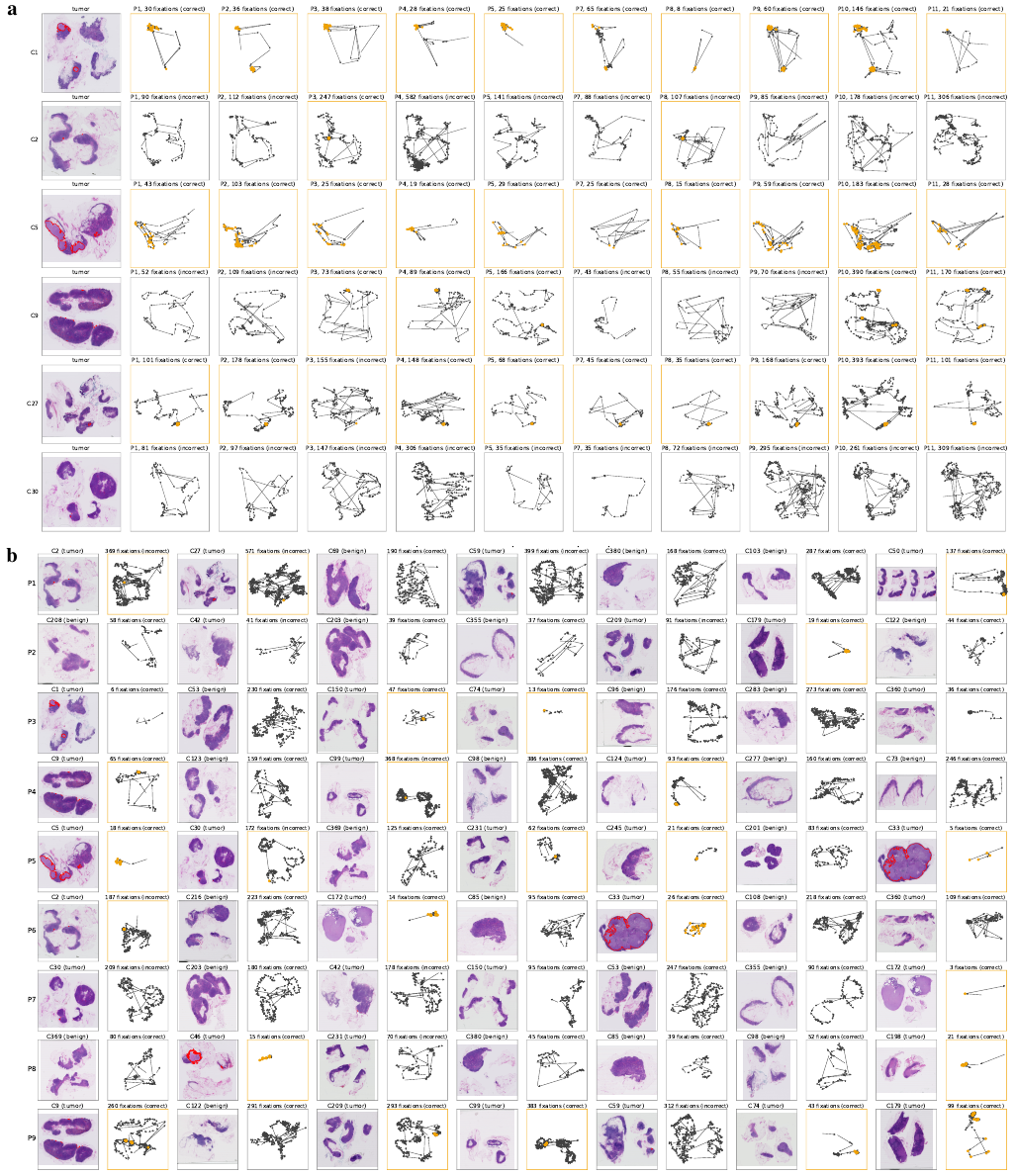}
    
    \caption{\textbf{Selected gaze of pathologists' behaviors in \dataname}
%Each WSI contains billions of pixels.
%Selected WSI viewing behaviors 
(a) in Experiment~I, in which multiple pathologists examined the same set of 60 slides, and 
(b)  in Experiment~II, in which no more than two pathologists viewed each slide. Tumor ground-truth regions are outlined in red\wordimg{Figures/icons/gtcir.pdf}, saccades are shown as dark lines\wordimg{Figures/icons/arrowb.pdf}, and fixation positions falling within tumor regions are in orange\wordimg{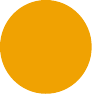}.
%shown as 
%large
%orange dots.
Images with orange borders\wordimg{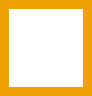} mark cases in which fixation positions fell inside tumor regions.  
Participant diagnostic outcomes are shown above each view. 
}
    \label{fig:E1E2Examples}
\end{figure}

\begin{figure}[!t]
    \centering
    \includegraphics[width=\textwidth]{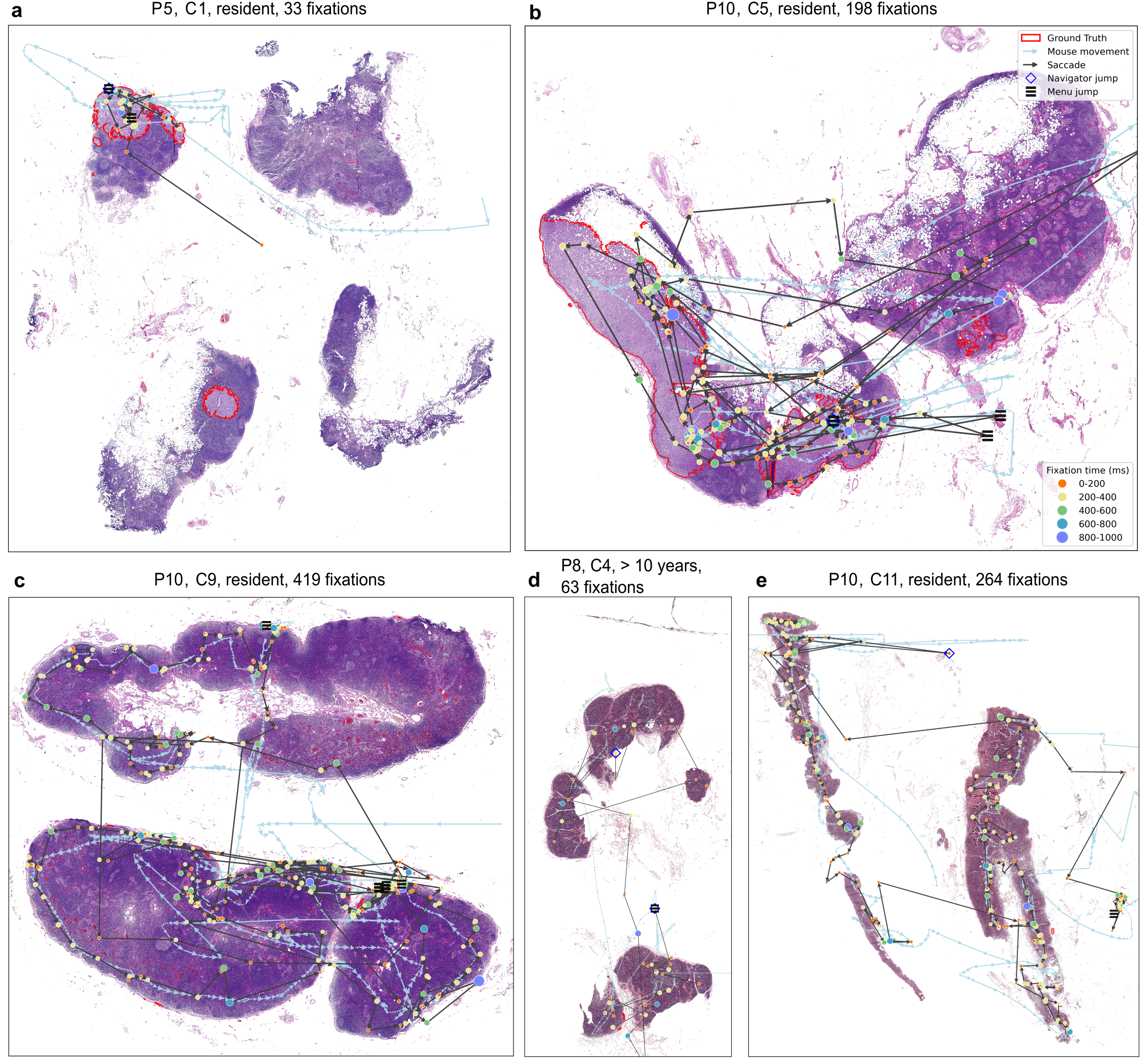}
\caption{\textbf{Selected examples of gaze and mouse behaviors from pathologists in \dataname.} 
%Each whole-slide image (WSI) contains billions of pixels (gigapixel scale). 
Overlaid on the WSI images are the ground-truth tumor tissue regions marked in red\wordimg{Figures/icons/gtcir.pdf}, 
fixations shown as dots whose sizes and colors encode fixation duration, saccades connecting successive fixations shown as black arrows\wordimg{Figures/icons/arrowb.pdf},
and mouse trajectories shown as light-blue arrows\wordimg{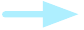}. 
Viewers could also navigate using an embedded navigator view. The blue diamond~\scalebox{1.6}{\wordimg{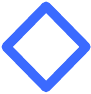}} represents these navigator jumps (viewing from the main view to the thumbnail navigator and vice versa). Fixations on the interface menu are represented by a horizontal-bar symbol\wordimg{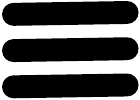} for menu jump (viewing from the main view to the right-pane menu and vice versa). 
Viewers could also navigate using an embedded navigator view. The blue diamond~\scalebox{1.6}{\wordimg{Figures/icons/dia.pdf}} represents these navigator jumps (viewing from the main view to the thumbnail navigator and vice versa). Fixations on the interface menu are represented by a horizontal-bar symbol\wordimg{Figures/icons/navi.pdf} for menu jump (viewing from the main view to the right-pane menu and vice versa). 
For cases with macrometastases, (a) participant P5 viewing slide C1 (2.29 mm diameter metastasis) made a diagnostic decision without fixating on all tumor areas. (b) P10 viewing C5 (11.26 mm) fixated on most of the tumor and tissue on their slide before responding. 
In micrometastases cases, (c, e) P10 correctly identified a tumor in C9 (0.81 mm) but incorrectly diagnosed C11 (0.28 mm). (d) P8 did not have any fixations on the tumor regions in C4 and incorrectly diagnosed the slide (0.27 mm).}
\label{fig:gazeExamples}
\end{figure}

\begin{figure}[!t]
    \centering
    \includegraphics[width=\textwidth]{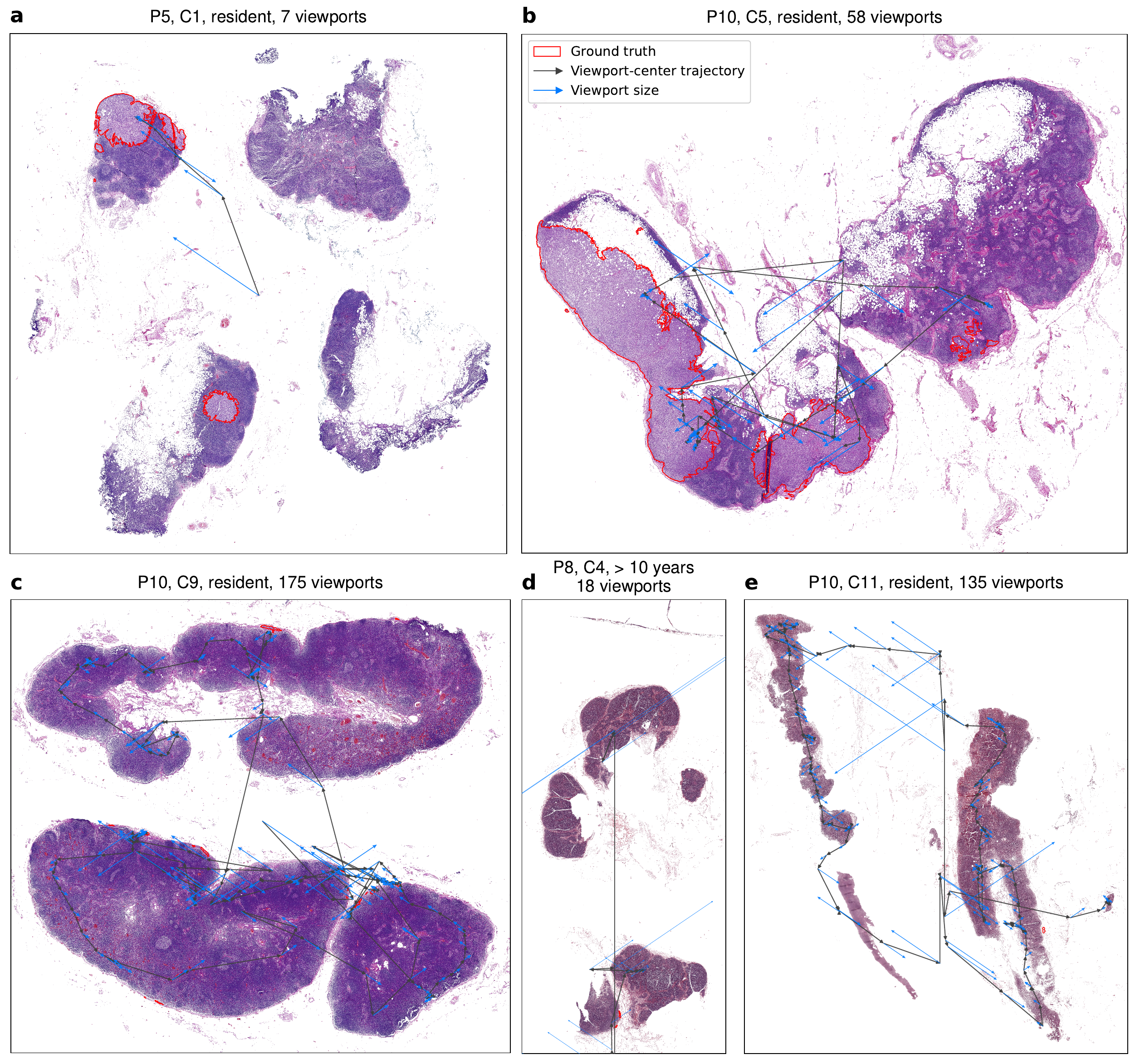}
    \caption{\textbf{Selected examples of viewport navigation in \dataname.} Overlaid on the selected regions of WSI images are the ground-truth tumor regions outlined in red\wordimg{Figures/icons/gtcir.pdf}, %the viewports shown as gray rectangles\wordimg{Figures/icons/view.pdf}, 
    and lines joining the centers of successive viewports shown as black arrows\wordimg{Figures/icons/arrowb.pdf}. %Each gray box shows the monitor viewing area; the smaller the box, the larger the zoom magnification.
    Each blue arrow\wordimg{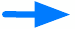} represents the monitor viewing area, with its length half the size of 
    %proportional to 
    the viewport diagonal; the shorter the arrow, the larger the zoom magnification.
    The blue arrow points toward the viewport's corner best aligned with the viewport's next movement. Participant P5 locates a region of interest using 7 distinct viewports and without using high levels of magnification. In contrast, P10, across cases C5 (58 viewports), C9 (175 viewports), and C11 (135 viewport), tends to navigate with higher zoom levels.}
\label{fig:viewportExamples}
\end{figure}

\begin{figure}[!t]
    \centering
    
    \includegraphics[width=\linewidth]{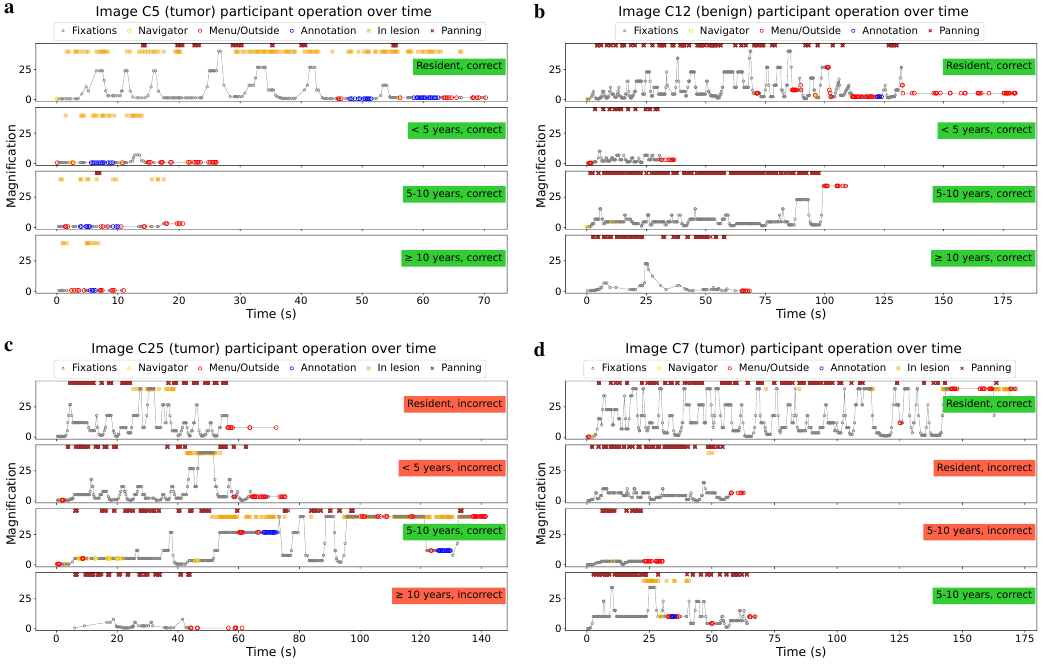}
    
    \caption{\textbf{Participants navigation behavior.} Each set of four plots shows four participants' navigation behavior while
viewing the same slide. Magnification is plotted against time, with circles
marking fixations during slide viewing (gray), navigator viewing (yellow),
menu/outside viewing (red), and annotation (blue). Orange squares
mark the times of fixations within tumor ground-truth regions
(``In lesion''), while brown crosses along the top of each plot indicate
panning. In-panel labels show participants' experience and diagnosis
correctness (green: correct; red: incorrect).
%\textbf{Observations.} In these examples, residents generally exhibited more frequent zoom changes and higher magnifications, whereas participants with $> 10$ years of experience tended to show fewer zoom changes and lower magnifications. 
The four cases include (a) an easy tumor case,
(b) an easy benign case, (c) a hard tumor case, and (d) a mid-range
tumor case.
}
    \label{fig:zoomcurves}
\end{figure}

\rv{The goal in this work is 
%therefore
to release structured data \rvvt{on pathologists' search behavior collected in} two experiments. First, we designed and implemented a testbed
\testbedname~(Fig.~\ref{fig:studySetupCoords}) 
%was designed and implemented 
to collect
%in naturalistic
%settings,  
gaze (without chin-rest), mouse tracking, and viewport tracking data, \rvvt{while}
%that mimics routine image navigation, 
instructing participants to search as they would during routine diagnostic tasks. 
This 
%type of daily use 
high-fidelity testbed-based method
can provide a scalable framework for data collection.
%understanding behavior and inconsistencies.
We  %further
computed a set of derived variables, for the community to have consistent data resources for downstream analysis or to supervise AI models.}
%but only if the data are 
% carefully captured and curated these data into structured representations. 
%In the process of understanding search behaviors for designing human-AI systems, we observed that pathologists spent about one to two minutes completing a task. This brief decision time led us to believe that the stimuli under the gaze point may contain important signals that could be used to behavior understanding, or supervise AIs and, conversely, to guide pathologists in the future, since pathologists must know \textit{where} and \textit{at what scale} to look first to finish tasks quickly. Given that annotated data are challenging to acquire, we could scale up our study potentially  collect routine fixation and navigation behaviors to study whether where pathologists look, for how long, what they miss, and why, that can assist pathologists in recovering from errors.
\rv{The primary contribution of this paper is \dataname~\cite{osfdataset}, which we collected using our testbed %. \dataname~contains data from 
from two experiments~(Fig.~\ref{fig:E1E2Examples}), including eye-gaze and mouse tracking~(Fig.~\ref{fig:gazeExamples}), viewport tracking~(Fig.~\ref{fig:viewportExamples}) and corresponding zoom and scanning~(Fig.~\ref{fig:zoomcurves}), and diagnostic-decision data for detecting breast cancer metastases in sentinel lymph node WSIs from CAMELYON16~\cite{bejnordi2017diagnostic}.}
% a broadly used public dataset released under \cczerolink.
% \rv{First, the captured, noisy data was converted into clean, AI-ready behavioral data in which multiple actions from multiple data-acquisition devices are synchronized.
%such as over- or under-interpretation (Fig.~\ref{fig:studySetupCoords}).
%(\autoref{tab:behaviors}).
% Second,
% \dataname~includes a set of computed variables, such as tumor size and area. Viewers’ objective behavior variables were labeled: \textit{viewing activities}, such as detecting particular regions of interest or recognizing their relevance; \textit{reasoning activities}, such as describing features or seeking information; and \textit{diagnostic outcomes}. These variables and activities can help determine the sampling scope when using \dataname.
% Finally, the data collection was validated by checking the quality of data synthesis.}
\rv{%With this work, 
Here, our data support 
%release the data 
%release can 
%to 
%addressed 
%the  central bottleneck
%in enabling  the community 
%to 
%in 
%validating
%the
behavior consistency %analyses
by running these two experiments 12 months apart, inviting the same pathologists back, and capturing 
%studies 
%-grounded errors 
%and 
behaviors in the same environment, creating a high-fidelity, clinically trustworthy dataset for intra- and inter-consistency analyses. 
%for behavior analyses and downstream  AI supervision.
We 
%have
preregistered the experiments 
%are %additionally preregistered 
and released all materials and data on Open Science Framework (OSF)~\cite{osfdataset}.
}

\section*{\scolor{Methods}}

\rv{This section reports on these two data-collection experiments and the design of a custom testbed, \testbedname. The data-collection goal was to support analyses of behavioral inconsistency and diagnostic errors, as well as AI supervision.
%To cross-validate our methods, we had many pathologists view the same slide set in the first experiment and increased the number of slides in the second experiment to cover the full set of source breast cancer images.
Experiment I captured where pathologists with different levels of expertise look and \rvvt{how they} reason by presenting the same 60 WSIs to all 10 pathologists; %Experiment II captured a large set of behaviors, providing robust multimodal data for training and evaluating pathologists and AI systems.
Experiment II 
%instead 
prioritized  coverage of a large WSI set of \totalwsicount  %CAMELYON16 breast cancer lymph node 
WSIs, broadening the range of behaviors captured. \rvvt{Here,} 9 pathologists each \rvvt{viewed} a separate subset.}
%. \rvvt{Each} of the 
%covering 
%\rvvt{was viewed by \totaletwopathologists pathologists}.

\subsection*{\sscolor{Whole-Slide Images and Readers}}

\rv{%The first data collection task was to select the images for behavior tracking. 
We selected the WSIs to meet the following criteria: (1) the dataset must be large enough to contain a diverse set of stimuli, with \rvvt{a large enough quantity of images} that supports diverse behaviors across observers for downstream error characterization %inter-rater and intra-rater consistency analysis,
and %model behavior 
AI supervision.
(2) It must have fully annotated
tumorous 
%tile
regions for comparing different observers. Such data are rarely available because of inconsistency in pathologists’ diagnosis decisions, but these annotations are necessary for researchers to evaluate attentional and navigational behaviors and to analyze the causes of errors and inconsistency.
(3) Since search difficulty is related to the size of the search target (in this case, the tumor)~\cite{nuthmann2021effect}, large tumor areas (macrometastases, cancer deposits that measure $> 2$ millimeters (mm) in diameter) tend to be easy to find and require little navigation. Therefore, cases should include small tumors (micrometastases, tiny clusters of cancer cells measuring $\in (0.2, 2.0]$ mm in diameter).
We decided to use 
% breast cancer type
lymph node slides 
%was chosen 
because of its relatively simple tissue structures and the challenge of diagnosing lymph node metastasis.}

These criteria led to the selection of the 
CAMELYON16 breast cancer lymph node WSIs, a publicly available dataset under \cczerolink~\cite{ bejnordi2017diagnostic, CAMELYON17}. 
It consists of 399 hematoxylin and eosin ($H\&E$)-stained breast cancer lymph node WSIs (239 benign and 160 with metastases), with ground-truth tile-level tumor area annotations stored as image masks, and a considerable variation of tumor sizes. 
These WSIs were collected from patients at %multiple 
two medical centers in the Netherlands 
%(University Medical Center Utrecht and Radboud University Medical Center)
and were scanned in sufficient detail, at either 20$\times$ magnification (pixel size 0.243~$\mu$m $\times$ 0.243~$\mu$m) or 40$\times$ magnification (pixel size 0.226~$\mu$m $\times$ 0.226~$\mu$m). \rv{As a result, they vary in stain intensity and sample preparation protocols, reflecting the practical challenges that computational models face in generalization.
In addition to the reasons listed above, CAMELYON16 is among the most widely used benchmarks for computational algorithms in digital pathology, with a reported $63\%$ of studies on tumor lymph node metastasis using it as of 2024~\cite{budginaite2024computational}. Thus, capturing pathologists’ viewing behaviors on this widely used benchmark allows comparison of attentional mechanisms between pathologists and AI observers. }
%The CAMELYON16 dataset is publicly available online under \cczerolink~\cite{camelyon16dataset}.

\textbf{Readers:} Participants were pathologists from The Ohio State University Wexner Medical Center's Department of Pathology \rvvt{who we divided into} four different levels of experience (residents, $\leq5$ years, $(5, 10]$ years, and $>10$ years), as suggested by our collaborating pathologist experts. 
%The department is also special in that it is the first pathology department in the nation to use purely digital viewing for primary diagnoses. 
The first experiment consisted of 10 participants: three residents, two with $\leq5$ years of experience, two with $(5, 10]$ years of experience, and three with $>10$ years of experience. The second experiment consisted of nine participants, of whom eight were returning: two residents, two with $\leq5$ years of experience, two with $(5, 10]$ years
%of experience 
(one new participant), and three with $>10$ years of experience (one $(5,10]$ that moved into this category in the time between experiments). 

\subsection*{\scolor{Tumor Size,  Tissue Area, and Tumor Grading}}

\rv{For data sampling and validation purposes, we calculated tumor diameter \rvvt{(along the longest axis)} from the CAMELYON16 tumor masks for all 160 tumor slides and categorized cancer spread using clinically significant tumor-size thresholds: \textit{isolated tumor cells} (ITCs) typically refer to single cells or tiny clusters measuring 0.2~mm or less in diameter ($\leq$ 0.2~mm); \textit{micrometastases} are deposits measuring $\in$ (0.2, 2]~mm; and \textit{macrometastases} are deposits measuring more than 2~mm in diameter, indicating a larger area of involvement in the lymph node. ITCs are generally not considered true metastatic disease in staging. 8 ITC slides, 93 micrometastasis slides, and 58 macrometastasis slides were identified.}

%\rv{
% for downstream attentional-behavior analyses and AI input sampling, where background regions are often removed.
% In contrast, the tumor grading itself is relatively coarse, with only the binary labels benign and tumor. 
%Tissue masks overlaid on thumbnails were visually inspected for a representative subset of slides and confirmed that they captured the tissue regions without substantial background or artifacts.
%Because tissue area was used only to stratify slide assignment to balance tumor-tissue ratios for participants in Experiment II, minor over- or under-segmentation at tissue boundaries was unlikely to materially affect the downstream sampling. 
%}

\rv{Cancer staging uses tumor-size thresholds, based on clinical significance, to categorize how far cancer has spread in these slides.
%After consultation with 
Pathologists on the team
%we updated 
suggested a tumor grading scale matching categories used in clinical settings: ``benign'', ``atypical'', ``neoplastic'', and ``malignant'', 
%. These four categories are 
%implemented in the testbed, \testbedname, 
shown as radio-button options on the testbed.
%though the experimenter may choose to only provide a subset of options for their specific experiment. 
For downstream correctness labels, ``malignant’’ and ``Neoplastic''  were mapped to CAMELYON16’s ``tumor'' labels. The ``atypical'' responses were mapped to the original CAMELYON16 tissue type labels.}

\subsection*{\scolor{Experimental Design}}
\subsubsection*{\sscolor{Experiment I (\eonedataname) Data Selection, Tasks, and Participants.}}
 \begin{wrapfigure}{l}{0.12\textwidth}
  \begin{center}   \includegraphics[width=0.12\textwidth]{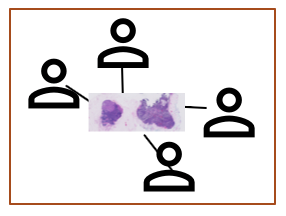}
  \end{center}
\end{wrapfigure}

\textbf{Design:} To facilitate study behavioral differences among pathologists, \rvvt{our first experiment, (\eonedataname), involved \totaleonepathologists} pathologists \rvvt{(P10)} who examined \rvvt{the same 60 (S10)} breast cancer lymph node slides drawn from CAMELYON16 --- 15 benign and 45 malignant (3 ITCs, 25 micrometastases, and 17 macrometastases) --- presented in random order during the experiment\rvvt{, a total of \totaleonetrialcount viewings (T600)}.

\textbf{Tasks:} As shown in~Fig.~\ref{fig:studySetupCoords},
participants performed a cancerous-tissue search task with three subtasks:
(1) label the cancerous tissue regions that led to their diagnosis, using the pen tool in \testbedname, (2) rate the tissue using the tumor grading scale, and
(3) briefly describe their rationale in a text box within the \testbedname~interface.
There was no time limit for completing the visual search and these decision tasks.

\textbf{Apparatus:}
The computer display was a 27'' Philips Barco monitor (MDPC-8127) with a resolution of 2,560 $\times$ 1,440, the same type of monitor pathologists used in their routine diagnostic practice. A Tobii Pro Fusion eye tracker (120 Hz) was attached to the monitor and connected to the Tobii Pro software. It did not require a chin rest to capture gaze data.
A secondary monitor displayed the live eye-tracking feed so that \rv{the experimenter} could
monitor the recording process. This monitor was under the desk and was not visible to the participant.

\subsubsection*{\sscolor{Experiment II (\etwodataname) Data Selection, Tasks, and Participants.}} 

\begin{wrapfigure}{l}{0.12\textwidth}
  \begin{center}
\includegraphics[width=0.12\textwidth]{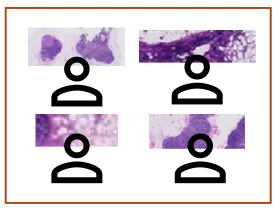}
  \end{center}
\end{wrapfigure}
\textbf{Design.}
%Given the  curation of the Experiment I dataset,
All 399 CAMELYON16 slides were used, with two reserved for training, to curate a relatively large behavioral dataset. \rvvt{\totaletwopathologists participants (P9)} received a unique sample of 60 slides \rvvt{from the 397 CAMELYON slides (D397), for a total of \totaletwotrialcount viewings (T540).} Each 60 slide sample comprised 36 benign slides randomly chosen from the CAMELYON16 dataset and 24 tumor slides sampled from the tumor slides using stratified simple random sampling based on the tumor-to-tissue ratio. 
Both sampling procedures were performed without replacement until all slides were sampled, upon which the sampling pool was refilled and sampling continued in the same manner, ensuring that each slide was viewed by \rvvt{at least one}
% one, and at most
% two
participant. 

\textbf{Task.} Participants only rated slides on a tumor scale in this experiment and ranked the slide difficulty on a scale from 1 (easy) to 5 (hard). 
% \rv{The tumor scale was simplified for this experiment, removing the ``neoplastic'' option}.
They were not asked to mark tumor regions, as Experiment I, 
in order to reduce the overall experimental time, and collect as \rvvt{much} high-fidelity behaviors as possible. 
%constrain the total experiment time and assemble a large dataset from their real-world clinical behavior, 
%with the goal of using eye behaviors to supervise machines. 
%The idea was that if pathologists viewed the slides for a limited time, their routine clinical behavior could be used to supervise machine-learning tools.
\rvvt{Participants were not asked to} input textual reasoning since \rvvt{the} notes \rvvt{from} Experiment I \rvvt{proved to be} too brief to be useful. We decided to skip that step and interviewed them instead.
They were still instructed to behave as they would in clinic diagnostic settings.
%naturally as possible.

\textbf{Apparatus.} All other apparatus details were the same as in Experiment I, except that eye tracking was performed at 60 Hz. There is no significant difference in the number of detected fixations between 60 and 120 Hz eye-trackers. There may be differences in detection rate and saccades~\cite{leube2017sampling}. \rv{For this reason, while the data can be compared within experiments, comparison across experiments should be limited to fixations, viewed regions of interests, scanpath, and overall diagnostic performance.}
%rather than saccade measures, which may not be comparable.

\subsection*{\scolor{Data Acquisition}}

\subsubsection*{\sscolor{Digital Pathology Application Grounded Data Collection Testbed (\testbedname)}}

\rv{We designed and implemented
a high-fidelity application-grounded testbed, \testbedname, 
%was
for collecting and recording behaviors: 
\textit{navigation} (viewport, mouse tracking, and keyboard events),  \textit{viewing} (eye-gaze), and \textit{diagnostic decisions}. This testbed-design approach can standardize a set of representative tasks and vary several factors at once (\eg, task difficulty, environment properties, and interaction techniques) while tracking multiple outcome measures simultaneously. 
It can also make new techniques %immediately 
comparable to conditions already run through the same testbed, since the diagnostic tasks and measures do not change between studies.
This experimental framework allows 
%the community
to control a broad range of variables consistently across studies.}

The testbed was implemented using the OpenSeadragon 3.1.0 library~\cite{OpenSeadragon} to load WSIs and support a broad range of interaction techniques. The interface design mimics a routine pathology workflow and interfaces with the Tobii Pro Lab software package to synchronize gaze-data collection. 
The testbed supports several interaction techniques:
%records the participant's mouse activities:
holding the left mouse button allows panning and zooming across the WSI, while the arrow keys allow navigation through the slide 
tile by tile, where a tile is sized at $224\times224$ pixels,
ensuring that no region of the tissue is overlooked. Second, it records mouse dragging to change the viewport, 
%as well as 
and a direct viewport placement through the navigator view in the `navigator overview'~(Fig.~\ref{fig:studySetupCoords}).
%upper-right corner of the screen. 
Finally, the annotation interface was implemented using Annotorious~\cite{Annotorious}, 
providing three shape options: a circle, a rectangle, and freehand drawing with the mouse. \rv{The green curve in Fig.~\ref{fig:studySetupCoords}}
illustrates an annotation made by a participant in the experiment. 
Participants could also edit their annotations by clicking to move or delete annotated line segments.
\rv{Diagnostic decisions were collected directly through the testbed interface.
% which shows up to four levels (``benign,'' ``atypical,'' ``neoplastic,'' and ``malignant''), expanding the binary ``benign'' and ``tumor'' ground-truth classifications from CAMELYON16. 
The first experiment, \eonedataname, used all four diagnosis options (``benign,'' ``atypical,'' ``neoplastic,'' and ``malignant''), while the second, \etwodataname, excluded the ``neoplastic'' category. 
% These more detailed categories were labeled by the pathologist on the team. 
% In \eonedataname only, participants typed notes explaining their decisions into a designated text box in the interface.
} 
During the experiments, \testbedname 
% supported these user interactions and 
logged all viewers' activities, such as mouse, screen viewport, slides' coordinates, magnifications, etc.

\subsubsection*{\sscolor{Experimental Protocol}}
Participants provided informed consent after receiving a briefing on the procedure. Demographic information was collected via online forms. Participants sat in front of the computer without a chin or head rest to mimic routine clinical use. Each participant was trained to use the testbed and follow the experimental procedure.
Their gaze was calibrated using a nine-point procedure in the Tobii Pro Lab software package. Fixations 
%and saccades
were computed using the Tobii I-VT Fixation Filter~\cite{olsen2012tobii} with the software’s default configurations. 
%\rvvt{While this filter reports both fixations and saccades, it is primarily designed for fixation detection. We report the fixations
%and saccades 
%as they are from the Tobii Pro Lab output, but the filter is highly sensitive to noise, such as instances where the eyes are not detected, or the velocity cannot be calculated. In these instances, the filter is prone to reporting multiple saccades without a fixation between, causing the saccade count to be higher than the fixation count.}
Participants in all runs passed the calibration tests before continuing. 
The eye tracker was calibrated for each participant and re-calibrated at the start of each recording session or \rvvt{on two occasions} when the experimenter \rvvt{visually} observed \rvvt{the screen recording displayed on the secondary monitor was not synchronized with the main monitor}.
Participants were instructed to behave exactly as they would when conducting a search in a clinical setting.
During Experiment I \eonedataname, they were told not to exhaustively
mark all tissue areas, but only those that influenced their final decision for tumor diagnosis. Participants were told that annotation was not needed for benign cases.
No other instructions were given. They practiced with two slides, a benign case and a micrometastatic case. They could ask questions during training and practice but were not allowed to ask questions during the formal experiment.
During the formal experiment, they viewed and completed the tasks for their assigned slides. Selecting a diagnostic decision from the \testbedname~interface was mandatory before they could proceed to the next slide, and they were not allowed to revisit previously viewed slides.
They took a mandatory two-minute break every 30 minutes. \rvvt{The eye-tracker was not re-calibrated for these breaks if the participant did not leave their seat.} 
After completing the viewing tasks, participants were interviewed about their subjective viewing experience and their attitudes toward eye tracking.

\subsection*{\scolor{Computational Data Processing}}

\rv{All variables are computed while pathologists perform search and decision tasks,} \ie, the period from the moment the WSI image finished loading to the time the participant clicked the ``Next'' button to proceed to the next reading.

\textbf{Synchronization between eye-tracking and WSI viewing behaviors.}
\rv{We faced the challenges Lopes et al. raised in eye tracking and WSI imaging, to associate and correlate real-time eye tracking data with real-time zoom levels  and the associated interpretation and analysis of the data~\cite{lopes2024eye}.}
%\todo{... summary sentence stating the variables we used in the synchronization step.}
%What we did was to
In our case, since both platforms, our testbed and the Tobii Pro Lab record time, 
we use these two timestamps to synchronize 
the 
eye-tracking 
data from the Tobii eye-tracker and all other tracking data (mouse, viewport, image, zoom, and panning) from the testbed 
%were synchronized 
for each task run.  
%so that each fixation recorded by Tobii Pro Lab was associated with the zoom, viewport, and image stimuli recorded by the testbed.
Tobii Pro Lab uses the Tobii I-VT filter to aggregate gaze samples and label them as single fixation 
%or saccade events 
and recorded the time of each fixation. 
The testbed \testbedname
%recorded 
also collected
timestamps for all events, 
%from the eye tracker, 
allowing synchronization between eye-tracker fixations and testbed events so that fixations could be associated with viewport position, image coordinates, and zoom level etc.
% The testbed and eye tracker recorded data at different frequencies 
%intervals,
%there were instances in which a fixation from the eye tracker occurred \textit{between} the testbed's data points. 
\rv{For  fixations where exact testbed data is not available, we 
can find the corresponding slide magnification, panning, and viewport using linear interpolations 
%used linear interpolation and forward-filled the testbed data (image width and height, zoom level \rvvt{(a continuous variable)}, image aspect ratio, and the viewport areas)
%(intermediate values used for later calculations of fixation position relative to the image])
%were interpolated 
between adjacent testbed events. We can then get the temporal scanpath trajectories. Lines between points are added to help visualize the paths and viewport transitions~(Figs ~\ref{fig:E1E2Examples},~\ref{fig:gazeExamples},~\ref{fig:viewportExamples}).} 
%data points, and calculated 
%relative timestamps 
%were calculated 
%for each reading task based upon the frame the image 
%was
%displayed on the screen.
%, and  peak saccade velocity
%was calculated  using adjacent gaze samples within each saccade. 
%The Tobii default setting of binocular tracking, averaging gaze position of the left and right eyes was used. 
%\end{comment}

All behavior modalities were %synchronized in
transformed 
%a common 
to the WSI image-coordinate system to facilitate use~(Fig.~\ref{fig:studySetupCoords}). 
Here, \textit{image coordinates} in the WSI image space defined by the pixel positions, with the origin $(0, 0)$ located at the upper-left corner and the coordinate range matching the WSI pixel resolution.
\textit{Screen coordinates}, by contrast, are used by the eye-tracking and mouse-tracking devices, with the origin located at the upper-left corner of the computer monitor and the data domain defined by the monitor resolution of $[2{,}560; 1{,}440]$ pixels.
The \textit{viewport coordinates} specify the boundaries of
the WSI region on the screen coordinates extracted from OpenSeadragon. Adjusting the viewport corresponds to panning or zooming operations that determine which portion of the WSI is displayed.
In \testbedname, shown in Fig.~\ref{fig:studySetupCoords}, the red rectangle shown in the \texttt{navigation window} represents the viewport in image coordinates.

\textbf{Mapping from screen space to image space.}
In addition to the raw eye-tracker outputs in screen coordinates, the transformed data in which fixation positions are mapped to image coordinates are included. 
This transformation is achieved by rescaling the screen-space values and adjusting for the current viewport offset relative to the screen and WSI. 
The coordinate conversion from screen position to image position was computed as follows: 
%\begin{equation}
\[
\begin{bmatrix}
x_{\mathrm{image}} \\
y_{\mathrm{image}}
\end{bmatrix}
=
\begin{bmatrix}
\frac{w_{\mathrm{viewport}}}{w_{\mathrm{window}}} & 0 \\
0 & \frac{h_{\mathrm{viewport}}}{h_{\mathrm{window}}}
\end{bmatrix}
\begin{bmatrix}
x_{\mathrm{screen}} \\
y_{\mathrm{screen}}
\end{bmatrix}
+
\begin{bmatrix}
x_{\mathrm{viewport}} \\
y_{\mathrm{viewport}}
\end{bmatrix}
,\]
%\end{equation}

%\noindent
where $(x_{\mathrm{image}},\, y_{\mathrm{image}})$, the image-space coordinates, 
$(x_{\mathrm{screen}},\, y_{\mathrm{screen}})$, the 
%raw 
screen-space coordinates, and $(x_{\mathrm{viewport}}, \, y_{\mathrm{viewport}})$, the viewport offset. 
$w_{\mathrm{viewport}}$ and $h_{\mathrm{viewport}}$, the viewport width and height reported by OpenSeadragon, while $w_{\mathrm{window}}$ and $h_{\mathrm{window}}$ correspond to the monitor width ($2{,}560$~px) and height ($1{,}440$~px), respectively.

\textbf{\rvjc{Zoom and scan levels}.}
\rvjc{Pathologists 
%make use of 
zoom in to change depth to change the  magnification of the slides to reveal features in higher detail for inspection or zoom out to obtain a birds-eye view of the slides to help determine the next region of interests. Our data and those of Drew et al.~\cite{drew2021more} show that the pathologist zooms in and out frequently to search for tumorous tissues.
% To facilitate the community to use the same set of metrics to study behaviors,
}
%we compute the magnification or absolute zoom levels ($m$), which is the slide size 
%in view relative to the full-resolution WSI:
%\begin{equation}
%\[
%\mathit{magnification} = \frac{w_{\mathrm{displayed}}}{w_{\mathrm{slide}}}. 
%\]
%\end{equation}

\rvjc{We followed Drew et al.~\cite{drew2021more} to define zoom as additive changes in magnification level between adjacent fixations: 
\[
zoom = \sum_{i=2}^{n} \left| \log_2\!\left( \frac{m_i}{m_{i-1}} \right) \right|,
\]
where $i$, the current fixation index ($i\geq1$), $n$, the total number of fixations on the image, and
$m_i$, the tissue magnification at the current fixation $i$. Doubling or halving magnification increments the zooming level by one, regardless of the magnification level at the previous fixation point.  We further compute \textit{Zooming rate}, which is \textit{zoom} normalized by the total interpretative time.
\textit{Scanning} describes the participant's viewing of multiple regions of the image to search for tumorous tissue, defined as the cumulative total distance of ``virtual'' eye-movements~\cite{drew2021more}. 
\[
scan = \frac{\sum_{i=2}^{n} s_i*\sqrt{(x_i-x_{i-1})^2 + (y_i - y_{i-1})^2}}{w} ,
\]
where $i$, the current fixation index, $n$,  the total number of fixations on the image, ($x_i$, $y_i$) is the coordinates of the $i^{th}$ fixation in image space, $s_i$ is the zoom scale for the $i^{th}$ fixation (scaled from the image space coordinates to screen space scale), and $w$ is the width of the viewing area (2,119 in our  experiments accounting for the menu where participants entered their responses). 
As a result, scanning accounts for fixations on different parts of the slide through scan path and through panning, even if the eyes do not move, such as during smooth pursuit. 
The \textit{scanning rate} is the scanning normalized by the total interpretive time.}
 
%\rvjc{}

 %between the time when the ``Next'' button is clicked and the time the next image is loaded.
%A random selection of image viewings was selected for manual verification. 
%The recorded fixations in the synchronized dataset were checked against the Tobii screen recording and its validation recordings to verify that fixations were in the correct positions and time and no additional fixations before and after each run appeared in the data.

\rv{\textbf{Inter- and Intra-rater consistency.}}
\rv{We designed the experiments to capture inter- and intra-rater consistency. 
We computed \textit{inter-rater consistency} among 10 pathologists evaluating
the same 60 WSIs in Experiment~I.
%to evaluate decision consistency. 
%\rv{Stemler~\cite{stemler2004comparison}, Gisev et al.~\cite{gisev2013interrater}, and Chaturvedi and Shweta~\cite{chaturvedi2015evaluation} suggested that consistency is a multifaceted concept that is fundamental to measurement science. It is either a primary or secondary component of pathology studies, in which pathologists serve as the measurement instrument.} 
We considered two commonly used measures of consistency: \textit{reliability} and \textit{consistency}. 
%Agreement between readers or intra-rater agreement within Experiment~I; 
%these comparisons concern consistency of the recorded diagnoses rather than accuracy against the reference labels.
\textit{Inter-rater reliability} describes how consistently readers
distinguish among items, whereas \textit{inter-rater consistency} describes
whether they assign the same rating to each WSI~\cite{gisev2013interrater}.
%\textit{Intra-rater consistency} is the degree of agreement among repeated evaluations or measurements performed by a single pathologist on the same WSI under similar conditions. Assessing this intra-rater consistencies requires data from repeated tests conducted at different times, which was collected in the two experiments 12 months apart. Participants overlapped between the two experiments.  Experiment I can measure inter-rater consistency and together with Experiment II intra-rater consistency.
%In Experiment~I, 
Here, we treated ratings (\eg, ``Benign'' , ``Atypical'', ``Neoplastic'', and ``Malignant'') 
%were treated 
as distinct categories, 
%and characterized consistency using the exact matching percentages.
and calculated 
%unweighted 
pairwise Cohen's $\kappa$ 
%for reader pairs to
for all 45 
reader pairs to  
%using their
%commonly rated slides,
assess the level of agreement between 
%two
pathologist raters, and  then averaged these as the final consistency. 
%with equal pair weights. 
%that is beyond random chance. 
We then used Fleiss' $\kappa$ 
%on completely rated slides as a supporting check.  Fleiss' $\kappa$ 
to assess the consistency 
%the same
for three or more raters~\cite{gisev2013interrater}.
%Pairwise Cohen's $\kappa$, and
%was 
%calculated for all 
%45 
%reader pairs using their
%commonly
%rated slides, 
%then averaged with equal pair weights. 
%We further
%computed 
%Complete-case 
%sensitivity analyses calculated %$\alpha$ and 
%computed Fleiss' $\kappa$ 
%on the 59 fully rated slides (590 ratings) 
%for assessing the reliability of agreements between these pathologist raters. 
%We calculated Cohen's $\kappa$ separately for each participant pairs and averaged the coefficients. 
Finally for 
%comparing
the \textit{agreement},
we calculated nominal Krippendorff's $\alpha$~\cite{krippendorff2004reliability} to compare individual ratings against 
%with  
that 
expected from 
%pooled category frequencies, including the finite-sample correction and 
all available ratings. 
We did not use 
Gallas et al.'s multi-reader multi-case
variance analysis of binary diagnostic performance~\cite{gallas2007binary}, because it quantified reader- and case-related uncertainty,
%in performance, 
whereas our analysis
%asks 
focused on whether the %original 
diagnoses align.}
%Our results are in the Technical Validation section and our computational details in the Supplementary Materials.
%, and is thus robust to missing data.

\rv{\textit{Intra-rater consistency} describes the extent to which a single 
pathologist assigns the same diagnosis to the same WSI on repeated
occasions. To assess this consistency, we matched 8 participants who took part
in both experiments, conducted 12 months apart, and compared their
recorded overlapping WSI diagnoses viewed in both experiments.}
% using exact-label matching
%percentages and chance-corrected coefficients.
% Mean within-slide exact consistency averaged the proportion of concordant
% reader pairs on each slide with equal slide weights. 

\rv{\textbf{Error classification.}
\rv{Diagnostic errors remain the most widely used metric, defined as the proportion of incorrectly classified samples among all cases.} To examine diagnostic errors, 
we also characterized navigation behavior~\cite{drew2021more}: here \textit{navigation} refers 
%simply 
to the pathologist physically moving the view, and a navigation error would be failing to bring a tumor region into the visible viewport at all, or the tissue not being displayed at a magnification sufficient to see it. 
Here, \textit{perceptual} errors are \textit{search} errors, in which the tumor was never fixated; \textit{recognition} errors %in which 
are errors in which the tumor was fixated only briefly and not long enough to register its significance); And  \textit{cognitive} errors occur at the decision level \ie, the tumor was fixated for an extended period, but the pathologist still reached an incorrect diagnostic conclusion.
Following Carmody et al.~\cite{Carmody}, we used 500~ms of fixation on the tumor %was used 
as the cutoff between recognition errors ($<$ 500~ms) and decision errors ($\geq$ 500~ms).
%Diagnostic accuracy maps for Experiments I and II are shown in~ and~Fig.~\ref{fig:interOnecase}, respectively. Together, they show how diagnostic outcomes vary across tumor-size groups and pathologist experience levels. 
}

\begin{figure}
    \centering
    \includegraphics[width=0.9\linewidth]{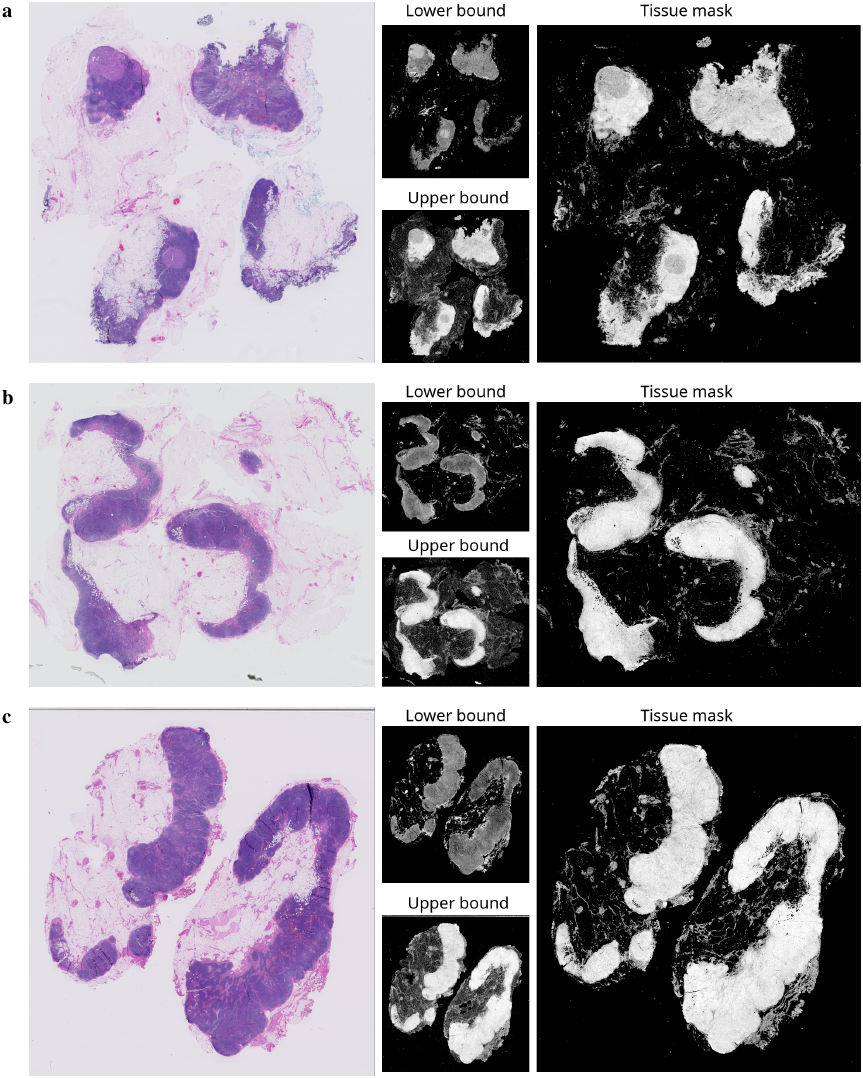}    
    
    \caption{\textbf{Otsu's tissue background segmentation results.} 
    %We provide several examples of how the tissue masks were created using HSV thresholding. We begin with the original images, in color.
   We used Otsu's method~\cite{otsu1975threshold}  to threshold the HSV values of the original images (left-column) and use these to create upper and lower color bounds for the image (mid-column). The lower bound recognizes areas of high saturation, and ignores darker areas, or shadows. The upper bound ignores bright areas, representing the white background. These bounds are combined (right-column) to create the final tissue mask.}
    \label{fig:otsu}
\end{figure}

\textbf{Foreground tissue and background non-tissue region classification.} For each WSI, We used Otsu's method~\cite{otsu1975threshold}, a standard method, 
%and 
%that 
to separate pixels into 
two classes of foreground (tissue) and background (non-tissue). 
%It returns a single intensity threshold  determined by minimizing intra-class intensity variance, or equivalently, by  maximizing inter-class variance.
We first analyzed the WSI's histogram of pixel intensities, 
%assuming 
of a bimodal (two-peaked) distribution, and then 
%calculating
calculated
the variance from all possible threshold splits to identify the best separating threshold. 
This process resulted in a binary image~(Fig.~\ref{fig:otsu}), and 
the lower bounds in the HSV color space, for the H and S channels 
%were  
automatically determined by Otsu’s method, while the lower bound for the V channel was fixed at 70 to retain well-saturated tissue regions and exclude shadows or dark areas.
We visually inspected the output. However, potential misidentification of tissue or background may arise from staining, scanning artifacts, and complex tissue morphology.

% For each WSI, tissue was selected using slide-specific HSV bounds of
% $\left[H_{\mathrm{Otsu}}, S_{\mathrm{Otsu}}, 70\right]$ to
% $\left[180, 255, V_{\mathrm{Otsu}}\right]$. The
% resulting Otsu thresholds ranged from 51--109 for hue, 21--89 for
% saturation, and 178--222 for the upper value bound.

\begin{figure}[!t]
    \centering
    \includegraphics[width=\textwidth, trim={0 0 60bp 0}, clip]{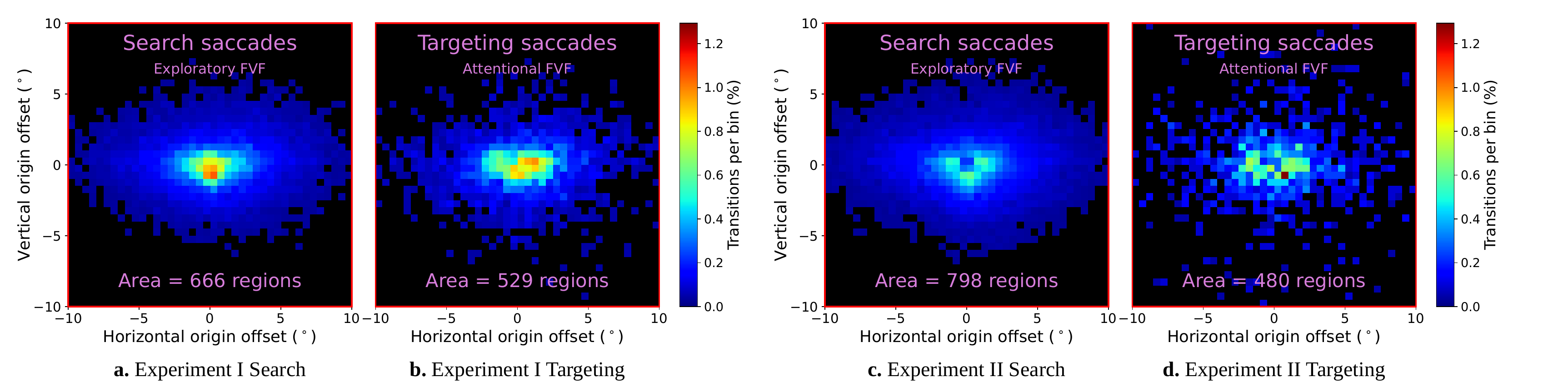}
    % \caption{\textbf{Functional visual fields.} The functional visual field is shown for Experiment I (a) and Experiment II (b), with saccade origins plotted relative to the corresponding endpoints for all participant readings. These reflect the regions where viewers can respond to visual stimuli and where they can plan eye movement. }
   \caption{\rv{\textbf{Exploratory (all slides) and attentional (tumorous slides only) functional visual fields.}
Results are shown for Experiment I (a--b) and Experiment II (c--d), 
with saccade origins plotted relative to their endpoints.
All subfigures use the same color scale, and the set of all colored pixels shows the smallest area that includes 75\% of all saccades.
The exploratory FVF shows saccades that do not \rvvt{land on} the tumor ground truth, in contrast to the attentional FVF, which shows saccades that do.}}
\label{fig:fvf}
\end{figure}

\textbf{Derived behavior variables.}
\rv{We computed a set of derived behavior metrics
(Table~\ref{tab:metrics}) 
%to provide 
%consistent
%, easy-to-use outcomes 
for downstream analyses. Beyond commonly studied variables, we constructed exploratory functional visual field (FVF) maps, extending Wu and Wolfe~\cite{wu2022functional}, to test whether this variable, originally designed to capture simple search in static scenes, can inform pathologists' \textit{dynamic} search.
We expect this variable to be useful to the community for studying why observers \rvvt{might} miss information \rvvt{that is present} in the field of view (including both the fovea and peripheral vision) and for gauging whether an observer has ``looked at'' \rvvt{the information} not only \rvvt{in} the fovea but also \rvvt{in} the periphery, which is necessary to \textit{guide} attention~\cite{wu2022functional, vater2022peripheral}.
%FVF define the effective region around fixation from which information can support target detection and eye-movement planning during visual search.
We have computed two FVF variables: 
(1) The exploratory FVF describes eye movements answering ``where should I fixate next?'' and is computed from saccades
that do \textit{not} go to the target within the current viewport.
The attentional FVF describes ``what should I attend to during this fixation?'' and is computed from %saccades 
scanpath
that do go to the target within the viewport. This is
%slightly
adapted from the FVFs calculated by Wu and Wolfe~\cite{wu2022functional} for the small and static images, 
%Our data differs from theirs as ours is 
to our dynamic search task that requires navigation in a large image and closer inspection of the target to determine %whether it is indeed 
a lesion. 
Within each experiment, we pooled eligible 
%viewport 
%saccades 
scanpath 
across participants for the tumor slides to compute both the exploratory and attentional FVFs. For the benign slides, we computed the exploratory FVF when no targets exist. 
Fixation centers approximated 
saccade endpoints, and screen coordinates were converted to visual angles in the observer's viewing coordinates and the eye positions recorded by the Tobii eye tracker.
We plotted each 
%saccade's 
scanpath 
fixation origin relative to its end fixation point, placed at $(0^\circ,0^\circ)$, to visualize where the eyes moved \textit{from}. The resulting 2D histograms used $0.5^\circ \times 0.5^\circ$ bins, were normalized by the total number of useful 
%saccades 
scanpath
within each experiment, and were displayed on a shared percentage scale. 
Fig.~\ref{fig:fvf} shows the smallest set of globally ranked bins containing at least 75\% of transitions.
%displayed in a $\pm10^\circ$ window.
}
%Wu and Wolfe distinguished search, targeting, and post-target saccades~\cite{wu2022functional}.
%We did not make this distinction because a unique targeting fixation is difficult to define in diagnostic whole-slide viewing. Our maps therefore
%summarize pooled eye-movement behavior rather than a separately identified search phase. near zero displacement and a distribution extending farther horizontally than vertically. 
%These maps provide descriptive FVF proxies, not independent measurements of peripheral target-detection ability. Future analyses could examine how
%these distributions vary with magnification or viewing stage.

%\textbf{Otsu method for tissue and background classification.}

%We provided their definitions of 
%defines 
%each
%these variables, and explain the rationale for their inclusion.
%The metrics were obtained by searching the literature for eye-tracking studies and collecting the behavior measures found. 
%The cleaned data contain the diagnostic workflow for each WSI reading, from the time the image was loaded onto the screen to the time the participant marked it as complete and proceeded to the next image.

\begin{table}[t]
    \rv{
    \centering
    \scriptsize
    \begin{tabular}{p{0.15\linewidth} p{0.30\linewidth} p{0.30\linewidth} p{0.15\linewidth}}
        \toprule
        \textbf{Metric name} & \textbf{Definition} & \textbf{Rationale} & \textbf{Source} \\
        \midrule
        Total fixations & Number of fixations in the \rvvt{slide} & More fixations indicate lower efficiency & Poole and Ball~\cite{poole2006eye}\\
        Fixations on tumor & Number of fixations on the tumor & Indicative of noticeability & Poole and Ball~\cite{poole2006eye}\\
        Total fixation duration & Total time of all fixations for a \rvvt{slide} & Compares the amount of attention across AOIs or stimuli & Poole and Ball~\cite{poole2006eye}\\
        Fixation duration on tumor & Total time of all fixations on the tumor & \rvvt{Also} compares the amount of attention across AOIs or stimuli & Poole and Ball~\cite{poole2006eye}\\
        Fixation rate & Number of fixations per second & Indicative of cognitive demands & Liu et al.~\cite{liu2022assessing}\\
        Time to first target fixation & Time from start up to first fixation on an AOI & Indicative of attention capture or target salience & Bylinskii et al.~\cite{bylinskii2015eye}\\
        % Time to first zoom & Time from start up to first magnification action & Indicative of knowing where to go & \\
        \midrule
        Zooming & The additive sum of changes in zoom level after a $\log_2$ transform & Tracks the amount of change in zoom on a slide with equal weight to multiplicative changes (\eg, 1x to 2x is weighted the same as 5x to 10x) & Drew et al.~\cite{drew2021more} \\
        Scanning & The cumulative distance in pixels of ``virtual'' eye movements & Tracks the amount of tissue that has been viewed, accounting for panning & Drew et al.~\cite{drew2021more}\\
        Zoom rate & Zooming as defined above, divided by the total interpretive time in seconds & Rate of zooming per second & Drew et al.~\cite{drew2021more} \\
        Scan rate & Scanning as defined above, divided by the total interpretive time in seconds & A rate of scanning per second & Drew et al.~\cite{drew2021more}\\
   %     Saccade count & Total number of saccades & More saccades indicate more searching, related to mental workload and cognitive processes & Fritz et al.~\cite{fritz2014using}\\
    %    Saccade peak velocity & Maximum speed within a saccade (deg/s) & Related to physiological arousal, mental workload, or the predicted value of information & Bruny\'{e} et al. (2019)~\cite{brunye2019review}\\
   %     Saccade amplitude & Degrees of visual field covered by the saccade & Indicates meaningful load cues; higher amplitude indicates lower effort & Poole and Ball~\cite{poole2006eye}\\
    %    Saccade duration & Total duration of all saccades & Related to mental workload and cognitive processes & Fritz et al.~\cite{fritz2014using}\\
  %      Saccade rate & Number of saccades per second & Related to task load and mental difficulty & Nakayama, Takahashi, and Shimizu~\cite{nakayama2002act}\\
        Functional visual field & Region around the fixation position where viewers can respond to stimuli & Area where viewers can plan eye movement and detect targets during a visual search task & Wu and Wolfe~\cite{wu2022functional}\\
        \midrule
        Pupil diameter & Diameter of the pupil & Physiological arousal & Math\^ot et al.~\cite{mathot2018safe}\\
        \midrule
        Coverage & Image area covered by fixations & Thoroughness of search & Bylinskii et al.~\cite{bylinskii2015eye}\\
        Tumor coverage & Tumor area covered by fixations & Thoroughness of viewing the tumor & Bylinskii et al.~\cite{bylinskii2015eye}\\
        \midrule
        Search errors & False-negative cases in which the tumor was never fixated & Search effectiveness & Kundel, Nodine, and Carmody~\cite{kundel1978visual}\\
        Recognition errors & False-negative cases in which the tumor was only briefly fixated & Ability to recognize the importance of a certain AOI & Kundel, Nodine, and Carmody~\cite{kundel1978visual}\\
        Decision errors & False-negative cases in which the tumor was fixated for an extended period & Ability to correctly classify the tissue & Kundel, Nodine, and Carmody~\cite{kundel1978visual}\\
        Navigation errors & False-negative cases in which the tumor was never brought into the viewport at zoom & Ability to navigate around the slide and locate a tumor via viewport zooming and panning & Drew et al.~\cite{drew2021more}\\
        \midrule
        Inter-rater kappa & Statistical measure of how likely two different raters are to agree on the diagnosis for the same slide & Reliability between different raters & Hanby et al.~\cite{hanby1992inter}\\
        Intra-rater kappa & Statistical measure of how likely the same rater is to agree when viewing the same slide at a later time & Reliability of the same rater & Hanby et al.~\cite{hanby1992inter}\\
        \bottomrule
    \end{tabular}
    \caption{\textbf{Behavior variables.}  Variables collected from the testbed \testbedname include eye-tracking (\texttt{fixation-related}, \texttt{saccade-related}, \texttt{pupil}, \texttt{coverage}, \texttt{error}, and \texttt{consistency}), \rv{mouse tracking, and zoom and pan levels}. 
    We calculated additional variables for 
    %These are provided for ease of 
    downstream analyses.
    }
    \label{tab:metrics}}
\end{table}

%Technical Validation II:

\subsection*{\scolor{Ethics Statement}}
\rv{This study was approved by The Ohio State University Institutional Review Board (IRB), the Office of Responsible Research Practices ($\#$2022B0227), including the study protocol and the release of collected data. All participants were briefed before the study and provided written informed consent to participate and for their gaze data to be collected and released. Participants could withdraw from participation at any time. Participant privacy is protected through the use of pseudonymous IDs.}

\section*{\scolor{Data Records}}
\label{sec:datarecord}

\rv{The dataset \dataname is available at OSF~\cite{osfdataset} under the
\ccfourlink. To make it accessible and reusable, the checklist, metadata, collected raw variables, and derived statistics are provided.
%used for validation as listed in~Table~\ref{tab:metrics}.
}
\rv{Each experiment's data is stored under the same named folders, \eonedataname and \etwodataname.
}
Both use the same data organization and format. 
A \texttt{Readme} file is provided in the root directory, describing the structure and content of each data file. 
\rv{The \texttt{Experience.csv} file provides the experience level of participants.}
\rv{The \texttt{participantCorrespondence.csv} file provides the matching between participants across the two experiments.}
\rv{The \texttt{ImageMapping.csv} file provides the mapping between the CAMELYON16 filenames (\eg, test\_001) and the IDs used by the experiments (\eg, C1).}
\rv{The \texttt{ImageMetrics.csv} file contains data about the WSIs themselves, such as width, height, and tumor area.
Within the experiment data folders, 
the \texttt{GazeOnlyData.csv} file contains only gaze events (fixations and saccades), with each event in its own row and with raw and calibrated timing. Similarly, the \texttt{MouseOnlyData.csv} contains only mouse events (panning, zooming, \etc) with raw and calibrated timing. The \texttt{CombinedGazeMouseData.csv} combines these gaze and mouse events into a single file using calibrated timing.}
\rv{The raw data folder contains files directly produced by the Tobii Pro software without filtering or calibration. This follows the most recent open-science standards, allowing the community to develop and apply algorithms to improve computational solutions. Finally, trial-level eye-tracking metrics are in the \texttt{GazeMetrics.csv} file. In this paper, a trial denotes a single run of WSI reading. The file \texttt{TrialBehaviorMetrics.csv} provides a list of summary behavior metrics for the dataset, including the specific measures and data points collected, derived, summarized, and recorded as metadata during the experiments.}

\begin{table}[!t]
\rv{
    \centering
    \scriptsize
    \setlength{\tabcolsep}{5pt}
%% Rotated version
    % \rotatebox{90}{
\begin{tabular}{c c c c c } 
% \rv{
\toprule
    \textbf{participantID} & \textbf{Validation Accuracy ($\degree$)} & Accuracy (px) & \textbf{Validation Precision ($\degree$)} & \textbf{Data Loss (\%)} \\
    \midrule
    \multicolumn{4}{c}{\textbf{Experiment I (E1)}}\\
    \midrule
E1.P1 & 0.30 & 17 & 0.21 & 0.18   \\
\rowcolor{gray!20}
E1.P2 & 0.31 & 16 & 0.27 & 0.03  \\
E1.P3  & 0.61 & 31 & 0.72 & 0.01  \\
\rowcolor{gray!20}
  & 1.51 & 83 & 0.99 &   \\
\rowcolor{gray!20}
\multirow{-2}*{E1.P4}  & $5.33^*$ & 252 & 1.19 &  \multirow{-2}*{0.33}   \\
 & 0.36 & 19 & 0.21 &   \\
\multirow{-2}*{E1.P5}  & 0.39 & 22 & 0.17 &  \multirow{-2}*{0.04} \\
\rowcolor{gray!20}
E1.P7 & $2.05^*$ & 89 & 0.39 & 2.13    \\
E1.P8  & 1.08 & 48 & 0.38 & 0.08  \\
\rowcolor{gray!20}
 & 0.59 & 32 & 0.51 &   \\
\rowcolor{gray!20}
\multirow{-2}*{E1.P9}  & 0.87 & 47 & 0.98 &  \multirow{-2}*{0.09} \\
E1.P10 & 0.54 & 25 & 0.73 & 0.003 \\
\rowcolor{gray!20}
& 0.70 & 39 & 0.16 &    \\
\rowcolor{gray!20}
 & 0.61 & 35 & 0.18 &    \\
\rowcolor{gray!20}
\multirow{-3}*{E1.P11} & 0.75 & 44 & 0.20 &  \multirow{-3}*{0.17} \\
\midrule
    \multicolumn{4}{c}{\textbf{Experiment II (E2)}}\\
    \midrule
E2.P1(=E1.P3)& 0.92 & 48 & 0.51 & 0.003 \\
\rowcolor{gray!20}
E2.P2(=E1.P1)  & 0.51 & 23 & 0.12 & 0.23  \\
 & $3.79^*$ & 181 & 3.03 &   \\
\multirow{-2}*{E2.P3(=E1.P4)} & 1.3 & 66 & 1.46 & \multirow{-2}*{0.09} \\
\rowcolor{gray!20}
E2.P4(=E1.P2) & 0.67 & 34 & 0.31 & 0.002 \\
E2.P5 & 1.48 & 75 & 0.53 & 0.58  \\
\rowcolor{gray!20}
E2.P6(=E1.P9) & $5.85^*$ & 382 & 0.99 & 0.45  \\
E2.P7(=E1.P11) & $2.14^*$ & 92 & 0.78 & 0.02  \\
\rowcolor{gray!20}
E2.P8(=E1.P8) & 1.06 & 59 & 0.36 & 0  \\
E2.P9(=E1.P10) & 1.03 & 51 & 0.83 & 0.008 \\
\bottomrule
\multicolumn{4}{l}{\footnotesize $^*$
Session exceeded recommended 
validation accuracy of ($>2\degree$).} \\
% }
    \end{tabular}
    \caption{\rv{\textbf{Experiment Eye-Tracker Quality Measures. }
    Per-participant measures of validation accuracy in degrees and pixels, validation precision using root-mean-square error in degrees, and data loss for both experiments. Each row represents a distinct recording session in which calibration was performed, and validation measures are reported for each session. Data loss is reported per participant and indicates the proportion of eye-tracker samples with no gaze information (neither eye detected). Data loss was below 1\% of gaze samples for all participants except one, whose data loss reached 2.13\%.
    %The vast majority of samples were usable (at least one eye detected). 
    \textbf{Observations:} Five of these experiment sessions exceeded $2\degree$ validation error. These errors most likely arose because participants were seated without a chin rest to preserve naturalistic viewing, or because tracking drifted over prolonged use.
    %All sessions passed the initial calibration. At our viewing distance, $1\degree$ corresponds to approximately xxx screen pixels.
    Fixation counts, fixation durations, and saccade measurement remain usable when drift occurs; fixation locations, however, are less reliable.
    }}
    \label{tab:measures1}
    }
\end{table}

\section*{\scolor{Data Overview}}
The dataset comprises \totalreadingcount WSI readings from \totaluniquepathologistparticipations~pathologists across the two experiments, with eight pathologists participating in both (Table~\ref{tab:measures1}), covering \totaletwowsicount unique lymph node WSIs from CAMELYON16 and \totalhourcount hours of recorded behavior. In \eonedataname, 
\totaleonepathologists pathologists independently viewed the same \totaleonewsicount WSIs, yielding \totaleonetrialcount readings. 
The corresponding behavior recordings total \totaleonehourcount hours. The folder contains 
\totaleonefixations fixations, %\totaleonemouse mouse action 
%\rvvt{samples} 
 and mouse clicking, panning, zooming,  annotating activities, %\totaleoneviewports 
 viewport changes, along with annotated regions associated with pathologists' tumor ratings. \etwodataname is comprised of \totaletwohourcount hours of data from \totaletwopathologists pathologists. Together, they contributed 540 readings of CAMELYON16 breast cancer WSIs drawn from the dataset's \totaletwowsicount unique WSIs,
yielding \totaletwofixations fixations, and 
%\totaletwosaccades saccades, \totaletwomouse mouse actions (clicking, panning, zooming), and 
\totaletwoviewports viewport change events from input devices.

\textbf{AI-readiness.}
In accordance with the FAIR (Findable, Accessible, Interoperable, and Reusable) data principles~\cite{wilkinson2016fair}, we structured \dataname~
%is structured 
%and documented 
to maximize transparency, accessibility, and long-term usability, by 
%All data files are accompanied
%by 
providing 
metadata, standardized naming conventions, and consistent formatting to ensure interoperability across analysis platforms. 
We used Croissant~\cite{akhtar2024croissant}  data structure, a high-level format that combines metadata, resource file descriptions, data structure, and machine-learning semantics into a single file. 
In our data, each
behavior table also includes descriptive headers and variable definitions, facilitating reuse.
%by both computational researchers and domain experts. 
Persistent identifiers and public repository hosting further ensure the dataset’s discoverability and accessibility in alignment with open science standards. 
\rv{For the set of derived variables, we have documented and released source code for all computations; they are independent of each other and can be replaced when better computational solutions or image quality metrics become  available.}
%We used Otsu, a most popular and computational efficient solution, it may introduce segmentation errors due to variations in staining, scanning artifacts, as well as complex tissue morphology.
%\autoref{fig:scans} shows a representative scanpath,
%is shown in~, 
%illustrating how gaze behavior, zoom level, and viewport navigation are recorded together.

\section*{\scolor{Technical Validation}}

\begin{figure}[!t]
    \centering

    \includegraphics[trim={5 0 0 50}, clip,
        width=0.99\textwidth
    ]{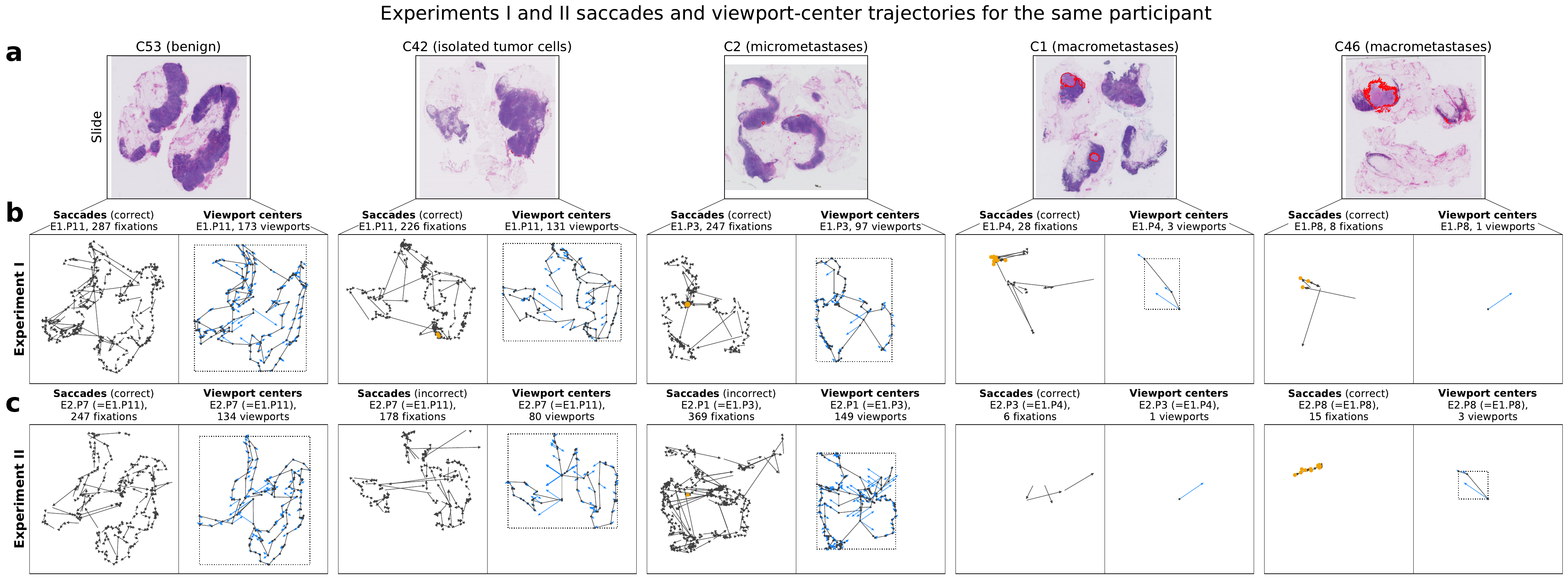}

\caption{\textbf{Comparison of gaze behavior between Experiments I and II in \dataname.}
Each column shows the same WSI and participant across the two experiments.
(a) WSI images, (b) Experiment~I gaze and viewport behavior, and
(c) the corresponding Experiment~II behavior. Red contours\wordimg{Figures/icons/gtcir.pdf} in (a) mark tumor ground-truth
regions, and large orange dots\wordimg{Figures/icons/orangecircle.pdf} in (b) and (c) indicate fixations falling within these
regions. Black arrows\wordimg{Figures/icons/arrowb.pdf} in the saccade sub-figures connect successive fixations to show saccade paths.
In the viewport sub-figures, dotted boxes\wordimg{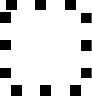} show viewport regions, black lines show viewport-center
trajectories, and blue arrows\wordimg{Figures/icons/arrowblue.pdf} indicate viewport movement. Panel titles
report participant IDs, fixation or viewport counts, and diagnosis status.
The examples include benign, micrometastatic, and macrometastatic WSIs,
including cases in which the participant's diagnosis differed between
the two experiments.}

    \label{fig:gazeExamplesE1E2}
\end{figure}

\rv{We 
%performed three types of validation 
%were performed
%to assess the reliability and accuracy of the multimodal dataset: 
validated eye-tracker calibration, compared our 
diagnostic accuracy to those from the CAMELYON16 paper, inter-rater and intra-rater consistency to those published in literature~\cite{hanby1992inter, riggi2026lobular, abdel2022inter}, and computed navigation behavior following Drew et al.~\cite{drew2021more}.
%The matched-participant comparison is shown in~\autoref{fig:gazeExamplesE1E2}, allowing gaze and viewport trajectories to be compared across repeated observations.
}

% \subsection*{\sscolor{On Calibration and Validation of Eye-tracking Data}}
% \section*{\scolor{Technical Validation}}
\subsection*{\sscolor{Eye-Tracker Quality Measures}}

\rv{
%Tracking calibration accuracy was validated in degrees of visual field and pixels using this dataset for both eye trackers: 120 Hz in the first experiment and 60 Hz in the second. 
% The sampling rate was not the same between the two experiments. Research has shown that there is no significant difference in number of detected fixations, but there may be differences in fixation duration and saccade detection rate and duration between 60 and 120 Hz eye-trackers~\cite{leube2017sampling}. Thus, comparison between experiments is best focused on measures such as the number of detected fixations, viewed areas, and overall diagnostic outcome, rather than saccade-related measures, as these may be unreliable. 
% Eye trackers typically operate at sampling frequencies from 25 to 250 Hz~\cite{andersson2010sampling} for capturing fixations and saccades. 
We calibrated the Tobii eye trackers 
%was calibrated
%during the experiments 
before each recording session and when drift occurred.
%as part of the data collection protocol.
%to ensure data quality. 
%Calibration occurred for each participant at the start of each recording session, excluding any of the %short 
%two minute breaks, 
%on the secondary monitor.
The data in the Tobii exports allow 
%users
us
to apply alternative validity and quality-control criteria. We reported the standard output, and summarized the %previous
aggregate estimates of recording precision in Table~\ref{tab:measures1}. 
%, and invalid-sample proportion were calculated from this data.
%In this paper, 
Data loss percent
%is reported 
is the percent of instances where the eye-tracker failed to record any eye data
%, \ie, where the eye-tracker failed to 
or detect either eye. 
Tobii treats all other instances, where one or both eyes are detected
%, are treated 
as valid samples.}
%during the viewing runs. 
%The only data removed are those in between viewing runs, before the next image is loaded and after the previous diagnosis is submitted.  
\rv{Across calibrations, mean validation error and precision (RMS) were 1.07$\degree$ and 0.49$\degree$ in the first experiment, \eonedataname (0.31$\%$ data loss) and 1.88$\degree$ and 0.89$\degree$ in the second \etwodataname (0.15$\%$ data loss).
Five sessions had larger validation errors. which were likely due to eye-tracking drifting over prolonged use. All sessions passed the initial calibration.
%At our viewing distance,
%$1\degree$ corresponds to approximately xxx screen pixels. 
Fixation counts and fixation durations remain usable when inaccuracy occurs; however, fixation locations could be less reliable.}

%\rv{For the first experiment, \eonedataname, the mean validation error was 1.07$\degree$ and the mean precision (RMS) was 0.49$\degree$ over all calibrations. The mean percentage of data loss was 0.31\%. For the second experiment \etwodataname, the mean validation error was 1.88$\degree$ and the mean precision was 0.89$\degree$ over all calibrations. The mean percentage of data loss was 0.15\%.}

%\rv{Immediately after each experiment, all recordings were manually checked against the screen recording to verify that the fixations in the collected and processed data matched those recorded directly by Tobii Pro Lab. Eye-tracking and mouse-tracking data were synchronized by aligning the timestamps recorded by the eye tracker in the Tobii data output with those logged by the testbed, allowing mouse events to be mapped to the eye-tracker's timestamps.}

% \rv{Gibaldi and Sabatini modeled the saccadic main sequence, a known relationship between saccade amplitude and other properties, such as duration and peak velocity~\cite{gibaldi2021saccade}. These predictable relationships can be used to model human saccades. The data was validated to determine whether the same main sequence relationships were observed in the experiments.}
% The same general trend was indeed observed: larger saccades exhibited higher velocities, leveling off at a saccade amplitude of approximately 10$\degree$. An exponential curve was fitted, consistent with the model reported to have the highest goodness of fit~\cite{gibaldi2021saccade}. 
% The resulting functional visual fields are shown in~Fig.~\ref{fig:fvf}.

\subsection*{\sscolor{Diagnostic Accuracy}}

\begin{figure}[!t]
    \centering
    \begin{subfigure}[T]{.95\textwidth}
        \includegraphics[width=\linewidth]{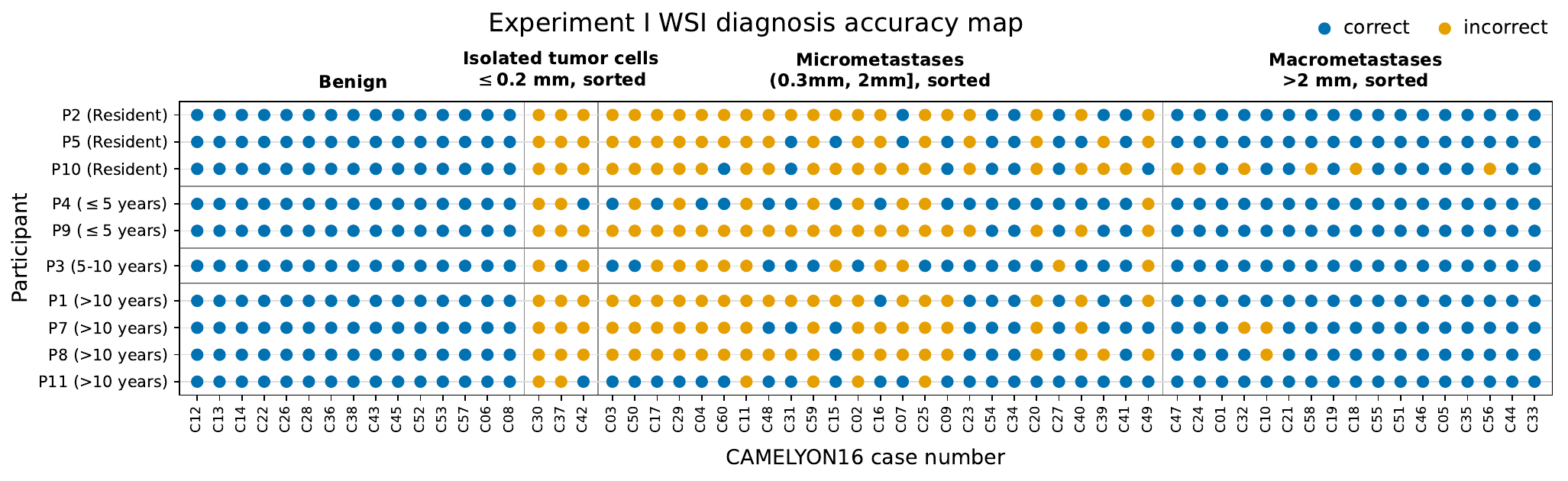}
        \end{subfigure}
    %\caption{\rv{\textbf{Diagnostic accuracy (Experiment I).} Diagnostic accuracy map for the 60 CAMELYON16 breast-cancer lymph-node WSIs viewed by all Experiment I pathologists. Rows are participants, grouped by clinical experience (resident; $\leq$5 years; 5–10 years; $>$10 years); columns are the shared 60 cases, grouped by ground-truth tumor burden (benign; micrometastases, (0.2~mm, 2~mm]; macrometastases, $>2$~mm). Each dot indicates whether a participant's diagnosis for that case agreed with the CAMELYON16 ground truth (correct, blue) or not (incorrect, orange). \textbf{Observations.} (1) Accuracy was highest and most consistent for benign slides, with errors confined to a single participant. (2) Errors were most frequent for micrometastases, occurring across nearly every participant regardless of experience level. (3) Accuracy on macrometastases was generally high.}}
    \caption{\rv{\textbf{Diagnostic accuracy (\eonedataname).} Each row represents one  participant, grouped by clinical experience (resident; $\leq$5 years; (5, 10] years; $>$10 years); columns represent the shared 60 cases, grouped by ground-truth lesion category (benign; isolated tumor cells (ITCs), micrometastases, macrometastases). Each dot indicates whether a participant's diagnosis for that case agreed with the CAMELYON16 ground truth (correct, blue) or not (incorrect, orange). \textbf{Observations.} (1) Accuracy was highest for benign slides. (2) ITC slides had errors for all but three participant--case judgments. (3) Micrometastases had a lower error rate than ITCs but contributed more errors because they comprised more slides than ITC. Errors occurred across every participant regardless of experience level. (4) Accuracy on macrometastases was higher than on ITCs and micrometastases.}}
    \label{fig:intraagreement}
\end{figure}

\begin{figure}[!t]
    \centering
    \begin{subfigure}[T]{\textwidth}   \includegraphics[width=\linewidth]{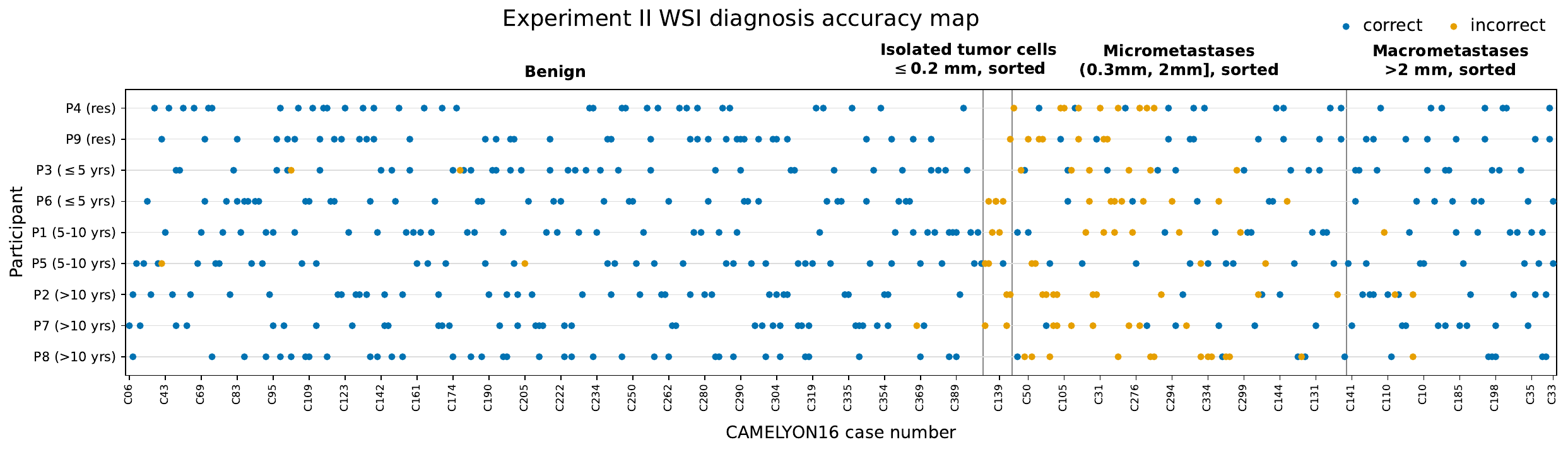}
        \end{subfigure}
        %\caption{\rv{\textbf{Diagnostic accuracy (Experiment II).} Diagnostic accuracy map for Experiment II, in which each of nine pathologists rated a unique sample of 60 CAMELYON16 slides (36 benign, 24 malignant) drawn without replacement from the full 397-slide set; because samples differ across participants, each row shows only the cases that participant viewed. Rows are participants, grouped by clinical experience (resident; $\leq$ 5 years; 5–10 years; >10 years); columns are cases grouped by ground-truth tumor burden (benign; micrometastases $\leq$ 2~mm, sorted; macrometastases $>2$~mm, sorted). Dot color indicates whether a participant's diagnosis agreed with the ground truth: correct (blue) or incorrect (orange). \textbf{Observations.} (1) As in Experiment I, benign slides were diagnosed correctly in almost all cases. (2) Micrometastases again showed the highest error rate, with most participants missing a substantial share of these cases. (3) Macrometastases were diagnosed correctly in most cases, with only scattered errors spread across a few participants, indicating that the size-dependent difficulty pattern from Experiment I held even with this larger slide set with fewer repeated slides. }}
        \caption{\rv{\textbf{Diagnostic accuracy (\etwodataname).} In \etwodataname, each of nine pathologists rated a unique random sample of 60 CAMELYON16 slides (36 benign, 24 malignant). Because samples differ across participants, each row shows only the cases that participant viewed. Rows are participants, grouped by clinical experience (resident; $\leq$5 years; (5, 10] years; $>$10 years); columns are cases grouped by ground-truth lesion category (benign; isolated tumor cells (ITCs), micrometastases, macrometastases). Dot color indicates whether a participant's diagnosis agreed with the ground truth: correct (blue) or incorrect (orange). \textbf{Observations.} (1) As in Experiment I, benign slides were diagnosed correctly in almost all cases. (2) ITCs had the highest error rate, with errors in 13 of 14 observed participant--case judgments. (3) Micrometastases had the next-highest error rate, with errors occurring across all nine participants and 67 of 132 observed participant--case judgments. (4) Macrometastases were diagnosed correctly in 71 of 75 observed judgments, with only scattered errors. 
        % Together, these observations indicate that the size-dependent difficulty pattern observed in Experiment I persisted in Experiment II, with the smallest tumor deposits presenting the greatest challenge.
        }}
    \label{fig:interOnecase}
\end{figure}

%Diagnostic accuracy maps for Experiments I and II 
%are shown in~
Figures~\ref{fig:intraagreement} and~\ref{fig:interOnecase} show diagnostic correctness for every observation,  
respectively, where blue dots show correct diagnosis and orange ones incorrect. 
%In , the common slide set reveals shared errors and inter-reader variation in correctness. 
\rv{
%Among the plotted Experiment~I,
%judgments, 
Accuracy was 100\% for benign
slides and 94.7\% for macrometastases compared with 10.0\% for ITCs and 41.2\% for micrometastases in \eonedataname.
%Errors on small tumor deposits occurred across experience groups, indicating that the difficulty was not confined to less experienced
% readers.
}
\rv{
%In 
Fig.~\ref{fig:interOnecase} showed that benign slides were again 
diagnosed correctly in almost all cases with 98.4\% accuracy, whereas
accuracy was 49.2\% for micrometastases and 94.7\% for macrometastases; ITCs remained the most frequently missed 
%lesion 
category.
%Together, these 
%results may support 
%a size-dependent difficulty pattern, where the smallest tumor deposits present the greatest errors and inconsistencies.
% These inconsistencies may not be attributed to experience.
}
%alone.

\rv{We also compared our participants' diagnostic performance in our Experiment II, \etwodataname to those reported in Bejnordi et al.~\cite{bejnordi2017diagnostic}.
%We compare only, since \eonedataname only used a subset of the CAMELYON16 slides. 
The binary participant classification yielded an area under the receiver operating characteristic curve (AUC) of 0.80, comparable to the AUC of 0.81 reported for pathologists evaluating CAMELYON16 slides under time constraints in Bejnordi et  al.~\cite{bejnordi2017diagnostic}, or 
at the lower boundary of the range conventionally described as excellent discrimination (AUC 0.80-0.90)~\cite{mandrekar2010receiver}.}
%Although our experiment imposed no time limit, 
%Our participants' 
%completed each slide 
%in approximately one minute on average, 
%approximately  the same time as Bejnordi et al.'s time constraint condition.

%The figures in the ``Diagnostic Accuracy'' subsection provide complementary visual evidence.
%Fig.~\ref{fig:matchedPairsFocus} illustrates stability and changes in correctness for the same reader--case combinations across experiments. These plots depict correctness rather than the original diagnostic categories and should therefore be interpreted alongside the exact-label agreement coefficients above.

%Correctness status differed between Experiments I and II in 10 of these 70 paired observations, while the remaining paired observations were either correct or incorrect in both experiments. The mismatches were distributed across lesion-size categories, with the greatest
%visual concentration 
%among the isolated tumor-cell and micrometastatic cases. This matched analysis complements the aggregate accuracy maps in Figures~\ref{fig:intraagreement} and~\ref{fig:interOnecase} by showing how diagnostic consistency varied for the same participant viewing the same case across experiments.

\begin{figure}[!t]
    \centering
    \includegraphics[
        width=0.98\textwidth
    ]{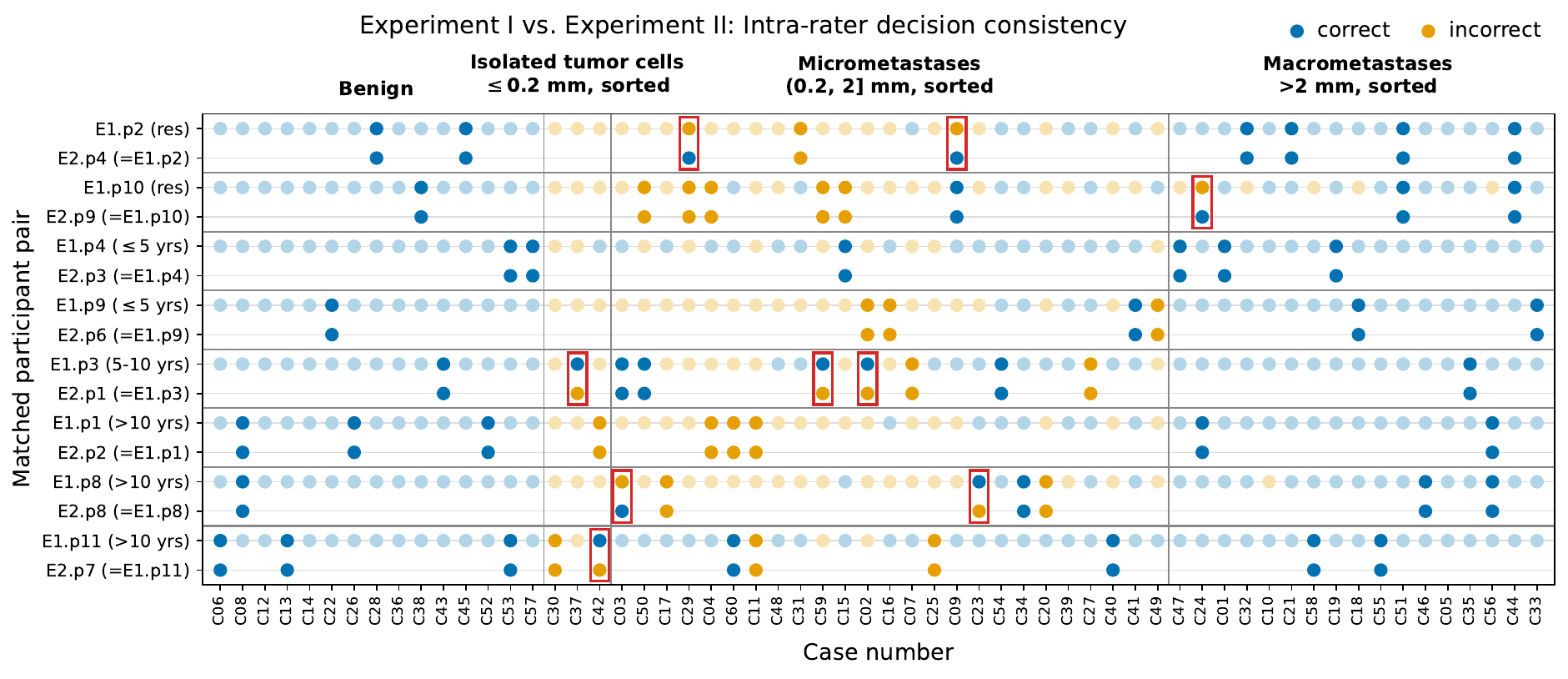}
    \caption{\rv{\textbf{Intra-rater consistency.}
    %The map  compare diagnostic decisions for the same cases between Experiments I and II  for eight matched participant pairs. 
    Each pair is shown as two rows, with the Experiment I observation in the top row and the same WSI diagnosis in Experiment II.
    %observation. 
    %Blue and orange dots indicate correct and incorrect diagnoses, respectively; faded dots indicate Experiment I observations without a corresponding Experiment II observation. 
    Red boxes highlight 
    %cases for which 
    inconsistent diagnosis cases in these two experiments.  
    %the correctness status differed between experiments. 
    %Cases are grouped by ground-truth lesion category: benign; isolated tumor cells, $\leq 0.2~\mathrm{mm}$; micrometastases, $(0.2~\mathrm{mm}, 2~\mathrm{mm}]$; macrometastases, $>2~\mathrm{mm}$.
    }
    }
    \label{fig:matchedPairsFocus}
\end{figure}

\subsection*{\sscolor{Consistency Validation}}
\rv{For Experiment~I, 
we assessed inter-rater consistency 
%was assessed 
by slide-level 
%matching
diagnostic
responses
%by slide 
and unweighted pairwise Cohen's $\kappa$
for the 45 pairs formed by the ten participants.
Across all 60 slides, the mean pairwise $\kappa$ was 0.555,
and the overall pairwise
%exact 
agreement 
%instead pooled reader-pair-slide comparisons:
was 2,045/2,700 
%overall
or 
75.7\%. 
%with 75.7\% pooled exact agreement. 
%Using the CAMELYON16 test reference-file groupings, 
To assess agreement jointly across all ten readers, we calculated  nominal
Krippendorff's $\alpha$ 
%was calculated 
across the 60 WSIs, obtaining 0.544 (95\%
slide-bootstrap CI 0.439--0.637). 
%Riggi et al. report a lower Krippendorff's $\alpha$ on of 0.3549~\cite{riggi2026lobular}, however, their work focuses on breast tissue and uses a different grading scale than this paper.
%In addition and as a supporting complete-case check, Fleiss' $\kappa$ was 0.539 across the 59 slides rated by all ten participants. 
Fleiss' $\kappa$ was 0.544 and fell within the ``moderate'' agreement range
(0.41--0.60)~\cite{landis1977measurement}, comparable to that of Abdel-Halim et al.'s %report a
Fleiss' $\kappa$ on lymph nodes of 0.61 and 0.68 across two rounds of testing, three months apart~\cite{abdel2022inter}.} 
This is slightly lower than the results reported by Hanby et al. on a different set of lymph node WSIs, with a mean inter-rater $\kappa$ of 0.67~\cite{hanby1992inter}.
One explanation could be tumor size---among the 397 CAMELYON16 WSIs,  101 contained micrometastases (93 slides) or isolated tumor cells (8 slides). We thus separated the consistency calculation by tumor size, and found that 
normal and macrometastatic slides yielded a mean pairwise $\kappa$ of 0.770 with 87.1\% exact agreement, whereas normal and 
micrometastatic slides yielded a mean pairwise $\kappa$ of 0.382 with 73.3\% exact agreement.
%However, $\alpha=0.544$ is below Krippendorff's suggested 0.667 threshold for tentative conclusions from coded data, indicating appreciable inter-reader variation rather than strong diagnostic consensus~\cite{krippendorff2004reliability}.
\rv{To assess intra-rater consistency across experimental settings, participants were matched across experiments \eonedataname and \etwodataname and compared their recorded
diagnoses on slides viewed in both experiments~(Fig.~\ref{fig:matchedPairsFocus}).
Eight participants contributed 70 matched reader--slide observations.
Unweighted Cohen's $\kappa$ was calculated separately for each participant
and averaged the eight coefficients, 
%with equal weight, 
obtaining 0.68;
diagnoses matched exactly in 57 of the 70 observations (81.4\%), 
%This is again slightly lower than Hanby et al.'s who report a mean intra-rater $\kappa$ of 0.79~\cite{hanby1992inter}.
%Under the same Landis--Koch convention, 
%this mean  falls within the  
or substantial agreement range~\cite{landis1977measurement}.}
%The estimate is based on small participant-level samples and is sensitive to differences in the recorded label sets: \eonedataname includes ``neoplastic,'' whereas \etwodataname does not. It therefore describes within-reader exact-label agreement across experimental settings rather than test--retest agreement under an identical protocol.}
%\rv{Computational details for both diagnostic-agreement analyses, including bootstrap uncertainty, complete-case sensitivity checks, and participant-level cross-experiment results, are provided in the Supplementary Materials.}

\section*{Usage Notes}

\rv{This dataset paper releases behavioral data that jointly capture navigational, perceptual, and cognitive behaviors during pathology diagnosis. 
%Compared with previous work that captured gaze data in controlled settings through explicit instructions~\cite{drew2023pathologist, mariam2022smart, drew2021more} or annotation tasks~\cite{mariam2022smart}, naturalistic data collection supports broader behavior understanding, including when and how AI behavior can align with, differ from, or complement pathologists' search strategies.
}
\rv{\dataname includes pathologists across a range of experience levels, from residents to those with more than a decade of experience, allowing the relationship between fixation onset and the development of visual expertise to be studied. In line with Simon's model of bounded rationality~\cite{simon1957models}, that humans use  heuristics rather than a full search process, our data may be used to examine how experts allocate attention under cognitive constraints, and how saccades and fixation, and fixation durations relate to diagnostic accuracy. %Pathologist competency could also be compared using the reliability measures described above. 
Understanding pathologists' behavior can %help the community characterize diagnostic errors and decision inconsistencies and 
support new approaches to supervising AI models or designing human-AI systems without imposing additional demands on their clinical time.}

\subsection*{\scolor{Informing Viewers' Strategies and Search Interface Design}}

\rv{Visual search on WSIs is extremely complex involving both central and peripheral vision, and yet pathologists are able to perform diagnosis and reason about the outcome within several years of training to become an expert. What computational principles can explain this extraordinary ability to capture the set of concepts that influence their behaviors? 
\dataname can help identify their strategies; 
regions where pathologists tend to struggle or exhibit lower accuracy, informing the small proportion of the regions pathologists looked at. 
%Because participants were instructed to behave as they would during routine clinical practice, 
\dataname
%~is well suited to studies of naturalistic error and inconsistency patterns. For example, the data 
can serve as a baseline for identifying diagnostic errors and for evaluating whether new tools designed to reduce them succeed in real-world workflows. 
Similarly, the data can establish a baseline for consistency measurement, both to assess whether two independent pathologists can be reliably trained to rate a given disease condition and to test whether new tools improve diagnostic reliability over time. 
We expanded upon the error taxonomy used in radiology (search errors, recognition, and decision errors)~\cite{kundel1978visual}.
%to Kundel et al.'s taxonomy of search errors, recognition, and decision errors. 
Even among expert pathologists, finding every tumor region is unnecessary and unlikely.
% This paper's hypothesis is that pathologists viewing under limited resources, humans must have learned to efficiently abstract their representations of the WSI.
}

\rv{One can ask whether, and for how long, a target region was fixated in the dynamic peripheral regions that facilitate search and by how much. The FVF data adds zoom level as an additional factor, allowing examination of whether patterns of panning and zooming are associated with visual field and errors. This could help characterize ``Looked But Failed To See’’ (LBFTS) errors~\cite{wolfe2022normal}, in which a pathologist fixates on a relevant region but fails to recognize it as important or targets within the attentional FVF when we count the peripheral vision~\cite{wu2022functional, vater2022peripheral}}
\rv{It can also improve the current design of next-generation digital pathology interfaces to better support human visual search strategies. For example, interfaces could highlight or warn about error-prone regions or provide navigation maps of areas likely to be missed. The first and last fixation in each trial, along with the corresponding diagnostic outcomes are available. These annotations can be used to relate exploratory scan paths and fixation durations to the location of the final fixation and the accuracy of the resulting diagnosis. Image quality metrics (IQMs) for unvisited regions could also help inform interface color design and support the study of image saliency. Such measurement remains inaccessible %or does not exist
for the cancerous tissue categorization. 
%Together, these measurement metrics could %provide
%a concrete pathway 
%be used for quantification and modeling performances, used in other domains, quantifying human-AI alignment and designing interfaces that accelerate residents' progression toward expertise.
}

%\subsection*{\scolor{Characterizing Errors and Inconsistencies in Real-World Diagnostic Workflows}}

% \rv{The working hypothesis is that errors in this data may stem from navigation failures as well as from perceptual or cognitive factors, the latter two of which have already been extensively studied in the literature. Because WSIs are stained, the visual, geometric, and textural properties of tissue structures may influence pathologists’ spatiotemporal reasoning during search. This current data do not separate navigation behavior from visual stimulus processing, and disentangling the two would require a new controlled study.}

\subsection*{\scolor{Human-AI Alignment and Digital Twins}}
\rv{\testbedname and \dataname could be used to build future human-AI systems that better integrate with and assist pathologists. Current systems tend to focus on high recall of tumor regions (emphasizing recognition of as much tumor as possible), but human pathologists do not examine all regions. 
Collected gaze data can be used to train models that predict where a pathologist will look, identify regions of interest that pathologists tend to see or miss, or actively guide a pathologist's attention to those regions in real time -- helping to avoid 
%search misses and LBFTS 
errors. 
Comparing the regions pathologists examine with those they skip could also help explain the causes of errors, though we note that pathologists need not examine every tumor region to reach a correct diagnosis. Moreover, because AI localization errors remain high, human feedback is likely to remain critical. Finally, 
this data also enables construction of a human attention map showing how experts view a slide, which can %serve as 
%ground truth
%to
evaluate whether AI saliency or attention maps align with expert visual reasoning, an approach relevant to explainable AI.}

\rv{\dataname can also support annotation-free settings for supervising AI algorithms, addressing the shortage of labeled training data by drawing on pathologists' natural viewing habits rather than requiring additional annotation effort. Predictive models trained on these data could further be used to build \textit{digital twins} -- virtual replicas of a pathologist's diagnostic process -- to simulate diagnostic reasoning for pre-screening slides, for stress-testing AI systems, or for educational purposes. With a larger dataset, digital twins could enable faster testing of human-AI interaction during algorithm and interface development without requiring constant pathologist involvement, support pre-screening to flag regions likely to need attention, and inform resident training by evaluating which interventions improve performance.}

%\begin{comment}
\subsection*{\scolor{Limitations}}
\rv{Although \dataname is, to the authors' knowledge, the largest public collection of pathologist behavior data of its kind, combining eye gaze, mouse, and viewport data, it has several limitations like any other data collections. 
Data come from 11 pathologists at a single institution, with uneven numbers of participants in the experience group. The two experiments also used two different eye-tracker sampling rate and some calibration sessions had larger validation error. 
We did not quantify the error of Otsu’s method in separating tissue from background, because the method has no analytic error bound: it selects a threshold from the image histogram. Validating it against manually labeled masks on a subset of slides is possible, but manual labeling is itself error-prone and impractical at this scale. We therefore performed only a visual inspection of the masks.
%the observed search strategies and diagnostic accuracy inevitably reflect the individual choices and experience levels of the participating pathologists at a single institution.
%The sample size of pathologists, while larger than prior work, may still be insufficient to fully generalize visual search behavior across tissue types. Our implementation of Otsu's method for tissue segmentation may limit tissue classification.} 
%the diversity and complexity of whole-slide images.
%Finally, pupil size and fixation duration can indicate emotional state, but there are cautions about associating the present data with fatigue: 
Our experiment did not %invite 
ask participants %did not 
to report fatigue.} %during the experiment.
%and the data may not account for confounding factors such as fatigue carried over from prior clinical work or insufficient viewing time to induce fatigue. Given the complexity of emotion-linked fatigue and these confounds, this data alone should not be used to draw conclusions about fatigue without additional validation.
%\end{comment}

\section*{\rv{Data Availability}}

\rv{The CAMELYON16 dataset is available under
\cczerolink. %and was used here with permission. 
% CAMELYON16 slide IDs, such as ``C1,'' ``C5,'' and ``C9'' in~Fig.~\ref{fig:gazeExamples}, were used to cross-reference collected eye-tracking data with the corresponding WSI images in all data tables.
} 
\rv{To make \dataname publicly downloadable, accessible, and reusable, all relevant information, such as the checklist, metadata, collected raw variables, and derived statistical data (for validation) in Table~\ref{tab:metrics} are uploaded to OSF~\cite{osfdataset} under
\href{https://creativecommons.org/licenses/by/4.0/}{Creative Commons Attribution 4.0 International License \mbox{\ccLogo\,\ccAttribution}}.}

\section*{Code Availability}

All data-processing code are released on OSF~\cite{osfdataset}. 
The code includes the eye-tracking and mouse-tracking event-alignment code, coordinate transforms, corresponding zoom and panning calculations, \rv{and metrics calculations}.

% \bibliography{bib}
% \include{bibliography}

\section*{Acknowledgments} 

The authors gratefully thank all the participants, professional pathologists, and residents who contributed to the empirical data collection.
The authors further thank Dr. Dayong Wang for discussions about CAMELYON16 regarding AI model training and sampling.
The authors acknowledge the Ohio Supercomputer Center for providing high-performance computing resources 
%and services 
that contributed to the research results.
%reported in this paper.
The work was carried out while R. Li, M. Ling, and S. Jiang were students at The Ohio State University, unrelated to their current roles at MathWorks, The Washington Post, and Google.

\section*{Funding}
This project was funded by the Ohio State University (OSU) Translational Data Analytics Institute (TDAI) seed fund. 

\section*{\rv{Author Contributions}}
\rv{All authors had full access to all the data and accept responsibility for submitting it for publication.
V.T., M.L., and J.C. drafted the initial manuscript. R.L. implemented the testbed, \testbedname, and configured the eye-tracking data collection environment, supervised by Z.L. and J.C., prepared the data, and collected the Experiment I dataset. 
R.L. and V.T. collected the two datasets and finalized the data cleaning.
M.L. prepared the image data for the empirical data collection and plotted all figures in the paper.
S.J. aided data collection.
J.C. and A.P. conceived the testbed and the gaze data collection idea.
J.C., Z.L., Y.H., A.P., J.W., and R.M. supervised students.
J.C. implemented data pre-processing, performed data cleaning and sanity checks, drafted the initial version, and finalized the manuscript.
All authors contributed to and approved the final manuscript.}

\section*{Competing interests} 

The authors declare no competing interests.

\clearpage 
\setcounter{table}{0}
\setcounter{figure}{0}
% \renewcommand{\tablename}{Table S}
% \renewcommand\thetable{\unskip\arabic{table}}
% \renewcommand{\figurename}{Figure S}
% \renewcommand\thefigure{\unskip\arabic{figure}}
%\section*{\scolor{Supplementary Methods}}

\section*{\scolor{Supplementary Material}}
%\subsection*{\scolor{Computational Data Processing}}

\subsection*{\scolor{Additional Data Synchronization Steps}}
We used the Tobii Pro Lab output with their I-VT filter to identify fixations and saccades. This  automatically categorizes gaze samples into fixations, saccades, eyes not found, and unclassified instances. They are then grouped with adjacent, matching samples, creating entire fixation and saccade events. Tobii labels each distinct fixation and saccade with a unique index. We retain this index in our data, so users may use it to refer back to the Tobii raw data. We dropped eyes not found and unclassified samples and grouped the Tobii data by event type (fixation or saccade) to create a list of single fixation or saccade instances. 

For the testbed data, we first used the \texttt{ExtractTestbedData.py} script to convert the raw testbed data stored as a JSON into CSV files for each participant. Next, \texttt{SynchronizeTestbedGaze.py} processed the timestamps in order to synchronize between the testbed and Tobii data. 
The testbed primarily uses a human readable timestamp and the Tobii data an integer clock timestamp. The testbed also recorded eye-tracker samples, including the Tobii clock. We converted the human-readable timestamps to an integer timestamp and used the first eye-tracker sample to adjust that timestamps to match. Now that the integer timestamps are synchronized, we merged the two data, so that the resulting output contains all data. For fixations or saccades that occurred during a testbed sample, we simply use that sample to assign navigation, viewport, and zoom information. It is possible to have a fixation or saccade event that did not occur during a testbed sample. In that case, we used linear interpolation to fill the entries. 

To create the final output files, we removed data in between the end of one trial and the start of the next, which are entries included in the testbed data. We also removed the testbed's recorded eye-tracker sample entries, as we have the fixation and saccade data from the Tobii data. We calculated and added a trial number column. At this point, the data file consists of the final combined gaze and mouse data. To create the mouse and gaze only data, we filtered out the gaze or mouse data.

\section*{\scolor{Data Records}}

\begin{figure}[!h]
    \centering
    \includegraphics[width=.6\textwidth]{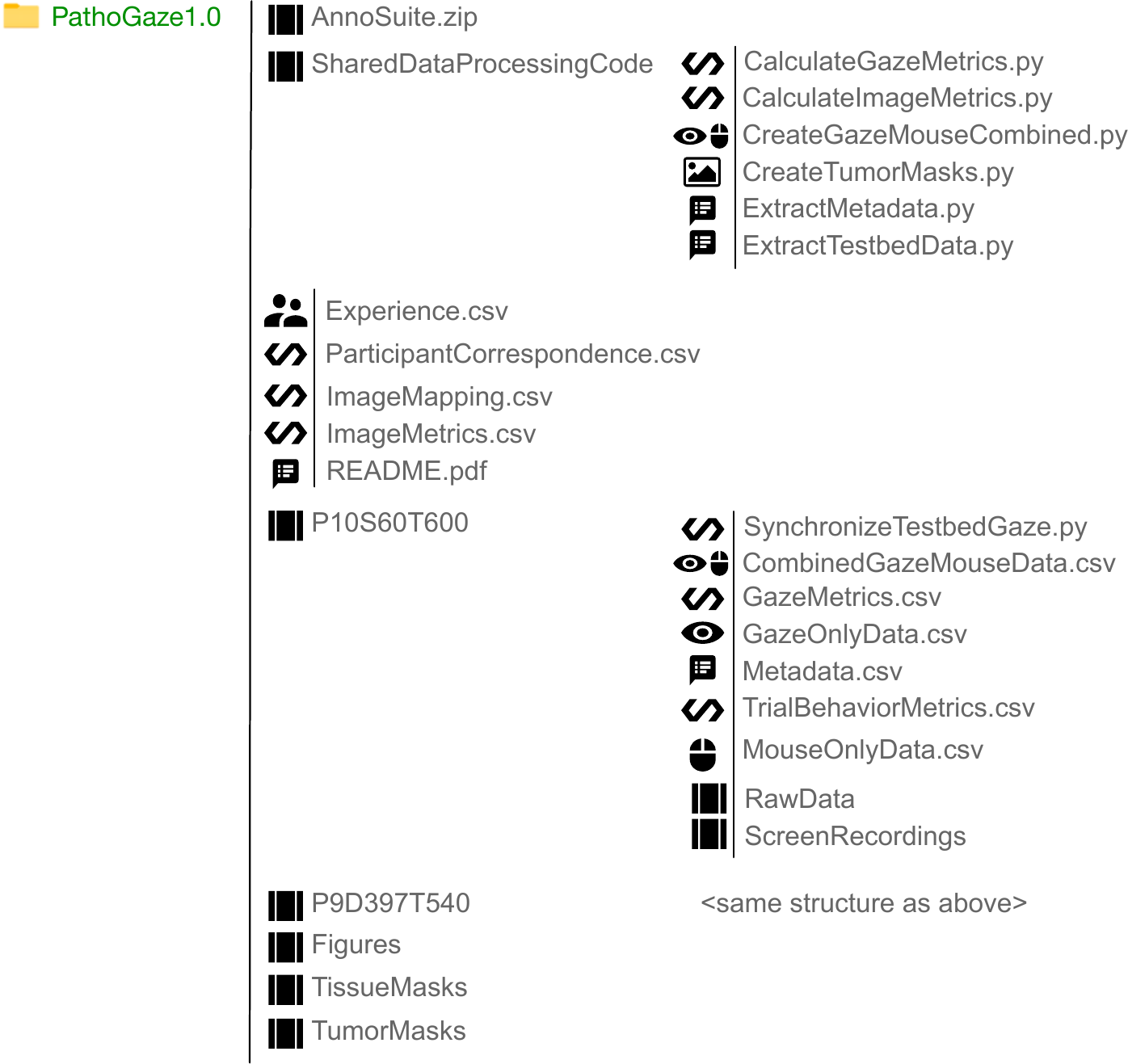}
    \caption{\textbf{Data directory structure.} %It places the common files, such as shared processing code, and image metric data at the root level.
    Shared resources such as common processing scripts and image metric data are placed at the root level. Experiment specific data are organized into subdirectories named after the first experiment, \eonedataname, and the second, \etwodataname. Both subdirectories contain the same type of data with the same organization: experiment specific code, raw testbed and eye-tracker data, screen recordings, and processed data.}
    \label{fig:folderstruct}
\end{figure}

\subsection*{\sscolor{WSI Image-based Variables}}

Located in the root directory (\texttt{ImageMetrics.csv}), this file consists of data regarding the images themselves. The columns are as follows: 
\textbf{A.} The \textbf{experimentImageName} is the name used by the CAMELYON16 dataset (test\_001, normal\_001, \etc). 
\textbf{B.} The \textbf{CAMELYONImageName} is a shortened ID identifying the slide used for the testbed. 
\textbf{C. width(px)} and \textbf{D. height(px)} are the width and height of the image in pixels, respectively.
\textbf{E. tissueArea(px)} and \textbf{F. nonTissueArea(px)} are the areas of the foreground and background, respectively, calculated by Otsu's method.
\textbf{G. diagnosis} is the ground truth diagnosis of the slide.
\textbf{H. tumorDiameter(mm)} is the diameter of the largest tumor region.
\textbf{I. tumorSize} denotes the slide-level classification based on the diameter of the tumor region: Benign, Isolated Tumor Cells, Micrometastasis, or Macrometastasis. Isolated Tumor Cells measure less than 0.2 mm in diameter, Micrometastases are greater than 0.2 mm but less than 2 mm, and Macrometastases are larger than 2 mm. 
\textbf{J. tumorMaskFile} is the link to the ground truth tumor mask for this image. 
\textbf{K. tumorMaskThumbnail} is a thumbnail image of the ground truth tumor mask.

The \textit{ImageMapping.csv} file contains the correspondence between CAMELYON16 file names and the IDs used by these experiments (\eg, test\_001 to C1). It additionally provides the lab that the slide was originally sampled from. It consists of three columns:
\textbf{A.} The \textbf{CAMELYONImageName} is the name from the CAMELYON16 dataset
\textbf{B.} The \textbf{experimentImageName} is a shortened ID identifying the slide used for the testbed. 
\textbf{C.} The {lab} is one of ``RUMC'' or ''UMCU'', indicating whether the slide was sampled at Radboud University Medical Center (RUMC) or University Medical Center Utrecht (UMCU). 

\subsection*{\sscolor{Participant Data}}
Located in the root directory, \texttt{Experience.csv} contains the experience level grouping of the participants. 
\textbf{A.} The \textbf{participantID} is a pseudonymous ID identifying the participant.
\textbf{B.} The \textbf{experience} is the experience level, one of these values: resident, $< 5$, (5, 10] years, or $> 10$ years experience, to indicate a resident, pathologist with less than 5 years of experience, 5--10 years, or more than 10 years of experience, respectively. 

Also in the root directory, \texttt{ParticipantCorrespondences.csv} contains the correspondence between participants across the two experiments. 
\textbf{A.} The \textbf{P10S60T600} is the participantID identifying the participant in the first experiment, P10S60T600. If this cell is blank, it indicates the participant in this row did not participate in this experiment.
\textbf{B.} The \textbf{P9D397T540} is the participantID identifying the participant in the first experiment, P9D397T540. If this cell is blank, it indicates the participant in this row did not participate in this experiment. 
\textbf{C.} The \textbf{experienceP10S60T600} is the experience level of the participant in the first experiment.
\textbf{C.} The \textbf{experienceP9D397T540} is the experience level of the participant in the second experiment.

\subsection*{\sscolor{Eye-, Mouse-, Viewport, WSI Region-Tracking, and Decision Dataset}}

\subsubsection*{Metadata}
\begin{table}[!t]
    \centering
    \scriptsize

%% Rotated version
    % \rotatebox{90}{
\begin{tabular}{c c c c c c c c c c c c c} \toprule
    \textbf{experimentImageName} & \textbf{participantID} & \textbf{trialID} & \textbf{taskCompletionTime(ms)} & \textbf{diagnosis} & \textbf{decisionNotes} & \textbf{annotations} & \textbf{correct} \\ \midrule
C48 & P1 & 1 & 36,139 & Benign & Benign & \{\} & FALSE\\
C4 & P1 & 2  & 37,117 & Benign & Benign & \{\} & FALSE\\
C58 & P1 & 3  & 65,717 & Malignant & Malignant & \{``Tumor":``(35,995.179... & TRUE\\
%\midrule
%C23 & P2 & 17  & 51,736 & Benign & No cluster... & \{\} & FALSE\\
%C54 & P2 & 18  & 53,538 & Malignant & Tumor mets... & \{``Tumor":``(18,551.708... & TRUE\\
%C13 & P2 & 19  & 59,413 & Benign & No cluster... & \{\} & TRUE\\
%\midrule
%C21 & P3 & 41  & 27,965 & Malignant & replacing ... & \%{``Tumor":``(71,962.539... & TRUE\%\
... & ... & ...  & ... & ... &  ... & ... & ...\\
%C41 & P3 & 42  & 15,876 & Malignant & replacing ... & \{``Tumor":``(33,726.613... & TRUE\\
%C35 & P3 & 43  & 14,954 & Malignant & replacing ... & \{``Tumor":``(43,919.996... & TRUE\\
%\midrule
%C47 & P7 & 11  & 38,167 & Malignant & malig & \{``Tumor":``(59,579.734... & TRUE\\
%\midrule
%C19 & P8 & 54  & 8,003 & Malignant & m & \{``Tumor":``(61,436.077... & TRUE\\
%\midrule
C14 & P9 & 5  & 456,833 & Benign & Dense fibr... & \{``Benign":``(40,382.11... & TRUE\\
    \bottomrule
    \end{tabular}
    % }
    \caption{\textbf{Excerpt of an Experiment I metadata file. } These files contain information about the trials themselves. Participants answers are provided in the diagnosis, decisionNotes, and annotations columns. The diagnosis is selected from the following options: ``Benign'', ``Atypical'', ``Neoplastic'', and ``Malignant''. Annotations are provided in JSON format, with an annotation type of ``Tumor'' or ``Benign'' and a list of the annotation coordinates in counterclockwise direction. Benign annotations are optional annotations where a pathologist may mark regions as benign. }
    \label{tab:meta1}
\end{table}

This file contains data from participants' trials, including summary information about their gaze data and trial-level attributes such as completion time and responses. This file is named \texttt{Metadata.csv} and appears under the \texttt{/\eonedataname} and \texttt{/\etwodataname} subdirectories. The columns are as follows:
\textbf{A.} The \textbf{experimentImageName} is a shortened ID identifying the slide used for the testbed. 
\textbf{B.} The {participantID} is a pseudonymous ID identifying the participant.
\textbf{C.} The \textbf{trialID} is the trial number for this participant. The first image a participant sees will have a \textbf{trialID} of 1, the second image a \textbf{trialID} of 2, and so on, up to 60. 
% \jc{fixation in benign, in the macro, and in the micro.}
\textbf{D.} The \textbf{taskCompletionTime} is the time in milliseconds for the participant to complete their tasks. 
% \jc{time in benign, isolated, in the macro, and in the micro.}
\textbf{E. diagnosis} records each participant's response to the diagnosis task, which can be Benign, Atypical, Neoplastic, or Malignant. 
\textbf{F. decisionNotes} contain notes provided by the pathologist explaining their diagnostic reasoning.
\textbf{G. annotations} list the pathologists' annotations in JSON format, with the annotation type as ``Benign'' or ``Tumor'' with a list of points in image-coordinates. 
\textbf{H. correct} represents whether the participants' diagnosis was correct or not, with TRUE representing correct, and FALSE incorrect.

\begin{table}[!t]
    \centering
    \scriptsize
    \setlength{\tabcolsep}{4pt}

%% Rotated version
    % \rotatebox{90}{
\begin{tabular}{c c c c c c c c c c c } \toprule
    \textbf{CAMELYONImageName} & \textbf{experimentImageName} & \textbf{width(px)} & \textbf{height(px)} & \textbf{tissueArea(px)} & \textbf{nonTissueArea(px)} & \textbf{diagnosis} & \textbf{tumorDiameter(mm)} & \textbf{tumorType} \\ \midrule
test\_001 & C1 & $8.60 \times 10^4$ & $8.96 \times 10^4$ & $8.24 \times 10^5$ & $7.71 \times 10^9$ & Malignant & 2.29 & Macro \\
test\_002 & C41 & $9.78 \times 10^4$ & $2.21 \times 10^5$ & $1.05 \times 10^6$ & $2.16 \times 10^{10}$ & Malignant & 1.34 & Micro \\
test\_004 & C42 & $9.83 \times 10^4$ & $1.04 \times 10^5$ & $1.25 \times 10^6$ & $1.02 \times 10^{10}$ & Malignant & 0.16 & Isolated \\
test\_014 & C43 & $1.11 \times 10^5$ & $1.00 \times 10^5$ & $1.50 \times 10^6$ & $1.11 \times 10^{10}$ & Benign & 0 & Benign \\
test\_053 & C12 & $9.78 \times 10^4$ & $2.19 \times 10^5$ & $3.00 \times 10^6$ & $2.14 \times 10^{10}$ & Benign & 0 & Benign \\
test\_099 & C30 & $9.83 \times 10^4$ & $8.60 \times 10^4$ & $1.32 \times 10^6$ & $8.45 \times 10^9$ & Malignant & 0.08 & Isolated \\
test\_105 & C33 & $1.27 \times 10^5$ & $9.32 \times 10^4$ & $4.13 \times 10^6$ & $1.18 \times 10^{10}$ & Malignant & 23.50 & Macro \\

    \bottomrule
    \end{tabular}
    % }
    \caption{\textbf{Excerpt of the WSI information.} 
    Images consisted of different diagnoses (benign and malignant) and tumor sizes.
    Slides were classified as isolated tumor cells ($\leq$0.2 mm), micrometastases ((0.2, 2] mm), or macrometastases (>2 mm) based on tumor diameter, while non-tumorous regions were labeled as benign.
    %classified into macrometastases (Macro), micrometastases (Micro), isolated tumor cells (Isolated) and benign (Benign), depending on the tumor diameter. Isolated tumor cells were less than 0.2mm in diameter, micrometastases were at least 0.2mm and at most 2mm in diameter, and macrometastases were larger than 2mm in diameter.
    %This file is located at xxx
    }
    \label{tab:im_metrics}
\end{table}

\subsubsection*{Eye-Tracking Data}
\begin{table}[!t]
    \centering
    \scriptsize
    \setlength{\tabcolsep}{3.5pt}

%% Rotated version
    % \rotatebox{90}{
\begin{tabular}{c c c c c c  } \toprule
    \textbf{experimentImageName} & \textbf{participantID} & \textbf{trialID} & \textbf{EyeMovementType} & \textbf{FixationPointX(px)} & \textbf{FixationPointY(px)}  \\
    \midrule
    C48 & P1 & 1 &  &  &  \\
    C48 & P1 & 1 & Saccade &  &  \\
    C48 & P1 & 1 & Fixation & 1,861 & 331 \\
    ...... & ...... & ...... & ...... & ...... & ...... \\

  %  \midrule
  %  C48 & P1 & 1 & Fixation & 2,341 & 68 \\
  %  C48 & P1 & 1 & Saccade &  &  \\
  %  C48 & P1 & 1 & Fixation & 1,195 & 495 \\
   % \midrule
   % C48 & P1 & 1 & Saccade &  &  \\
   % C48 & P1 & 1 & Fixation & 2,295 & 59 \\
    C48 & P1 & 1 &  &  &  \\
    \bottomrule
    %\\
    \multicolumn{6}{l}{\textit{Continued}}\\
    \toprule
    \textbf{ImageFixationPointX(px)} & \textbf{ImageFixationPointY(px)} & \textbf{EventEyesPositionX(mm)} & \textbf{EventEyesPositionY(mm)} & \textbf{EventEyesPositionZ(mm)} & \textbf{peakVelocity(deg/s)}\\
    \midrule
      &  &  &  &  & \\
      &  & 266.02 & 96.74 & 622.39 & 107.91\\
    171,996.33 & 50,832.50 & 267.83& 95.6& 624.58 & \\
    %\midrule
    %245,718.30 & 10,442.93 & 308.01&72.93 & 688.93 & \\
     % &  & 288.26 & 73.57& 683.79& 120.33\\
    %1,195 & 495 & 276.92 & 74.73 & 681.86 & \\
    %\midrule
    ...... & ...... & ...... & ...... & ...... & ...... \\
     % &  & 321.93 & 87.75& 663.05 & 73.38\\
    103,128.42 & 81,090.34 & 314.49 & 84.53& 669.18 & \\
    \bottomrule
   % \\    
    \multicolumn{6}{l}{\textit{Continued}}\\
    \toprule
     \textbf{ViewportUpperLeftX(px)} & \textbf{ViewportUpperLeftY(px)} & \textbf{ViewportWidth(px)} & \textbf{ViewportHeight(px)} & \textbf{relativeTimestamp(ms)} & \textbf{GazeEventDuration(ms)} \\ \midrule
        &  &  &  & 0 & \\
        -113,829.89 & 0 & 32,5451.78 & 221,184 & 148 & 58 \\
        -113,829.89  & 0 & 325,451.78 & 221,184 & 206 & 125 \\
        ...... & ...... & ...... & ...... & ...... & ...... \\

    %    \midrule
     %   -113,829.89  & 0 & 325,451.78 & 221,184 & 2,056 & 258 \\
      %  -113,829.89  & 0 & 325,451.78 & 221,184 & 2,314 & 75\\
       % -113,829.89  & 0 & 325,451.78 & 221,184 & 2,389 & 92\\
        %\midrule
        %33,542.656 & 79,301.60 & 64,249.34 & 43,708.77 & 35,822 & 42 \\
        %33,542.66  & 79,301.60 & 64,249.34 & 43,708.77 & 35,865 & 308 \\
         & &  &  & 36,139 &  \\

    \bottomrule
    %\\
        
    \multicolumn{6}{l}{\textit{Continued}}\\
    \toprule
      \textbf{outOfScreen} & \textbf{inMenu} & \textbf{inImage} & \textbf{inNavigator} & \textbf{zoomScale} & \textbf{zoomMagnification}\\ \midrule
      &  &  &  &  \\
      &  &  & 0.0065 & 1\\
    ...... & ...... & ...... & ...... & ...... & ...... \\
    % FALSE & FALSE & FALSE & FALSE & 0.0065 & 1\\
    %\midrule
    %FALSE & TRUE & FALSE & FALSE & 0.0065 & 1\\
    % & &  &  & 0.0065 & 1\\
    %FALSE & FALSE & TRUE & FALSE & 0.0065 & 1\\
    %\midrule
    % & &  &  & 0.033 & 5.062\\
    %FALSE & TRUE & FALSE & FALSE & 0.033 & 5.062\\
     & &  &  &  & \\   
     \bottomrule
    \end{tabular}
    % }
    \caption{\textbf{Excerpt of an eye-tracking data file. } The eye-tracking data files provide information for each fixation and saccade. Additionally, these data files have entries for the start and end of each trial for timing purposes. These entries 
    %do not 
    represent events associated with participants' fixations calculated in this study. 
    }
    \label{tab:eyetracking1}
\end{table}

This file contains recorded temporal gaze data along with viewport information. It includes all participants' fixations, saccades and
the viewports (corresponding visible areas of the WSI images). 
Each row in the gaze data file corresponds to a single event: either a fixation, a saccade, or the beginning or end of a trial. The trial start and end entries include values only for the imageID, participantID, trialID, and relativeTimestamp columns.
This file, named \texttt{GazeOnlyData.csv}, is located within the \texttt{/\eonedataname} and \texttt{/\etwodataname} subdirectories. The column names are: 
\textbf{A.} The \textbf{experimentImageName} is a shortened ID identifying the slide used for the testbed, ranging from ``C1'' to ``C397''. 
\textbf{B.} The {participantID} is a pseudonymous ID identifying the participant.
\textbf{C.} The \textbf{trialID} is the trial number for this participant. The first image a participant sees will have a \textbf{trialID} of 1, the second image a \textbf{trialID} of 2, and so on, up to 60. 
\textbf{D. EyeMovementType} indicates whether this row is a fixation, a saccade, or a start or end row. It takes the value ``Fixation'' for fixation rows and ``Saccade'' for saccade rows.
\textbf{E. FixationPointX(px)} and \textbf{F. FixationPointY(px)} are the x- and y-coordinates in pixels, respectively, of each fixation on the computer screen. %These columns are empty for the beginning of trial and end of trial rows.
\textbf{G. ImageFixationPointX(px)} and \textbf{H. ImageFixationPointY(px)} represent the x- and y-coordinates, respectively, of each fixation on the WSI image, measured in pixels. 
%These columns are empty for the beginning of trial and end of trial rows.
\textbf{I. EyesPositionX(mm), J. EyesPositionY(mm)}, and \textbf{K. EyesPositionZ(mm)} represent the average eye position during this event, in millimeters, with the upper-left corner of the screen representing (0, 0, 0). 
\textbf{L. peakVelocity(deg/s)} is the peak velocity of the saccade in degrees of visual field per second. This column only has a value for saccade rows. 
\textbf{M. ViewportUpperLeftX(px)} and \textbf{N. ViewportAreaUpperLeftY(px)} are the x- and y-coordinates of the upper left corner of the \testbedname viewport on the WSI image. 
\textbf{O. ViewportWidth(px)} and \textbf{P. ViewportHeight(px)} are the width and height of the viewport in teh WSI image space. 
\textbf{Q. relativeTimestamp(ms)} contains the time, in milliseconds, that the fixation occurred relative to the moment the image first appeared on the screen. \textbf{R. GazeEventDuration(ms)} is the fixation duration in milliseconds.
%This column is empty for the beginning of trial and end of trial rows.
\textbf{S. outOfScreen} is a TRUE or FALSE value indicating whether this fixation fell outside of the monitor screen. %This column is empty for the beginning of trial and end of trial rows.
\textbf{T. inMenu}, \textbf{U. inImage}, and \textbf{V. inNavigator} are TRUE or FALSE values indicating whether this fixation fell in the task menu, in the displayed image, and in the navigator window, respectively. 
\textbf{W. zoomScale} contains the zoom level of the image relative to the full size WSI image resolution.
\textbf{X. zoomMagnification} gives the zoom level of the image relative to the fully zoomed-out WSI image.

\subsubsection*{Action Data}

\begin{table}[t]
    \centering
    \scriptsize
%% Rotated version
    % \rotatebox{90}{
\begin{tabular}{c c c c c } \toprule
\textbf{experimentImageName} & \textbf{participantID} & \textbf{trialID} & \textbf{totalFixations} & \textbf{totalOrientingFixations}\\
    \midrule
    C48 & P1 & 1 & 100 & 15\\
    C4 & P1 & 2 & 111& 13\\
    C58 & P1 & 3 & 188 & 10\\
    ......\\
 %   \midrule
 %   C54 & P3 & 25 & 169 & 31\\
  %  C7 & P3 & 26 & 86  & 8\\
   % C6 & P3 & 27 & 146& 6 \\
    %\midrule
    %C15 & P11 & 8 & 96  & 1\\
    \bottomrule%\\
    \multicolumn{5}{l}{\textit{Continued}}\\
    \toprule
\textbf{totalFixationDuration(ms)} & \textbf{averageFixationDuration(ms)}  & \textbf{totalSaccades} & \textbf{totalSaccadeDuration(ms)} & \textbf{averageSaccadeAmplitude(deg)} \\
    \midrule
    28,742 & 287.42  & 110 & 3,914 & 5.76\\
    28,987 & 256.52  & 138 & 4,160 & 5.02\\
    52,125 & 277.26  & 214 & 7,388 & 5.40\\
    ......\\
%    \midrule
 %   47,521 & 281.19  & 223 &7,507 & 7.12 \\
  %  23,858 & 277.42 & 110 &3,933 & 7.21\\
   % 39,855 & 272.98 & 192 &6,582 & 5.94\\
   % \midrule
   % 21,216 & 221 & 243 &7,288 & 6.04\\
    \bottomrule\\

    \end{tabular}

    \caption{\textbf{Excerpt of an eye-tracking metrics file. } The eye-tracking metrics files provide trial-level eye-tracking metrics. %These are the metrics 
    %listed in \autoref{tab:metrics}, with the exception of saccade peak velocity, which is a saccade-level metric and is instead included in the eye-tracking data files. 
    %These metrics reflect participants' effort, mental workload, and cognitive processing. 
    }
    \label{tab:eye2}
\end{table}
\begin{table}[!t]
    \centering
    \scriptsize
    \setlength{\tabcolsep}{5pt}
%% Rotated version
    % \rotatebox{90}{
\begin{tabular}{c c c c c c } \toprule
    \textbf{experimentImageName} & \textbf{participantID} & \textbf{trialID} & \textbf{MousePositionX(px)} & \textbf{MousePositionY(px)} & \textbf{ImageMousePositionX(px)} \\
    \midrule
    C48 & P1 & 1 &  &  &  \\
    C48 & P1 & 1 & 1,301 & 718 & 85,990.4 \\
    C48 & P1 & 1 & 1,308 & 718 & 87,065.6  \\
    ......\\
  %  \midrule
  %  C48 & P1 & 1 & 966 & 1,016 & 78,892.67  \\
  %  \midrule
  %  C48 & P1 & 1 &  &  &   \\
 %   \midrule
  %  C48 & P1 & 1 & 2,114 & 855 & 97,640.30  \\
   % C48 & P1 & 1 & 2,114 & 856 & 97,640.30 \\
    C48 & P1 & 1 &  &  &   \\
    \bottomrule%\\

    \multicolumn{6}{l}{\textit{Continued}}\\
    \toprule
    \textbf{ImageMousePositionY(px)} & \textbf{ViewportUpperLeftX(px)} & \textbf{ViewportUpperLeftY(px)} & \textbf{ViewportWidth(px)} & \textbf{ViewportHeight(px)} &\textbf{relativeTimestamp(ms)}\\
    \midrule
    & -113,829.89 & 0 & 325,451.78 & 221,184 & 0\\
    110,284.8 & -113,829.89 & 0 & 325,451.78 & 221,184 & 288\\
    110,284.8 & -113,829.89 & 0 & 325,451.78 & 221,184 & 304\\
 ......\\
 %\midrule
    %110,167.04 & 70,214.66 & 101,009.32 & 19,069.44 & 12,907.29 & 25,058\\
    %\midrule
     %& 33,542.66 & 79,301.60 & 64,249.34 & 43,708.77 & 28,076 \\
    %\midrule
    %105,282.18 & 33,542.66 & 79,301.60 & 64,249.34 & 43,708.77 & 36,084\\
    %105,312.52 & 33,542.66 & 79,301.60 & 64,249.34 & 43,708.77 & 36,100\\
     & 33,542.66 & 79,301.60 & 64,249.34 & 43,708.77 & 36,139\\
    \bottomrule%\\

    \multicolumn{6}{l}{\textit{Continued}}\\
    \toprule
    \textbf{eventType} & \textbf{inMenu} & \textbf{inImage} & \textbf{inNavigator} & \textbf{zoomScale} & \textbf{zoomMagnification}\\ 
    \midrule
    image\_loaded &  &  &  & 0.01 & 0.96\\
    mouse\_move & FALSE & TRUE & FALSE & 0.01 & 1\\
    mouse\_move & FALSE & TRUE & FALSE & 0.01 & 1\\
    ......\\
   % \midrule
    %mouse\_scroll & FALSE & TRUE & FALSE & 0.11 & 17.09\\
    %\midrule
    %click\_button &  &  &  & 0.03 & 5.06\\
    %\midrule
    %mouse\_move & FALSE & TRUE & FALSE & 0.03 & 5.06\\
    %mouse\_move & FALSE & TRUE & FALSE & 0.03 & 5.06\\
    next\_btn\_clicked &  &  &  & 0.03 & 5.06\\
    \bottomrule
    \end{tabular}
    % }
    \caption{\textbf{Excerpt of a mouse-tracking data file. }
    These mouse events are temporally aligned to the eye-tracking data and are recoded 
    for the same start and end of each trial as the eye-gaze data. 
%An \textbf{eventType} column indicates the type of event, included image\_loaded for when the image is loaded, beginning the trial, and next\_btn\_clicked indicating when the participant clicked the button ending this trial. The click\_button event indicates when the participant clicks a menu button to select answers. 
    %These three event types do not track the mouse position, and so the corresponding columns are empty.
    }
    \label{tab:mousetracking1}
\end{table}
%This file consists of the mouse-tracking navigation data. This includes 
All participants' mouse movements and state of the 
% \testbedname 
viewports were recorded. Each row represents a distinct mouse interaction or the beginning/end of a trial.
This file, named \texttt{MouseOnlyData.csv}, is located in the folders \texttt{\eonedataname} and \texttt{\etwodataname} subdirectories. The column names are:
\textbf{A.} \textbf{experimentImageName} is a shortened ID identifying the slide used for the testbed, ranging from ``C1'' to ``C397''. 
\textbf{B.} {participantID} is a pseudonymous ID identifying the participant, ranging from ``P1'' to ``P10'' for the first experiment and ``P9'' for the second.
\textbf{C.} \textbf{trialID} is the trial number for this participant. The first image a participant sees will have a \textbf{trialID} of 1, the second image a \textbf{trialID} of 2, and so on, up to 60. 
\textbf{D. MousePositionX(px)} and \textbf{E. MousePositionY(px)} are the x- and y-coordinates of the mouse cursor, respectively, on the screen. %
%These columns are empty for the beginning of trial and end of trial rows.
\textbf{F. ImageMousePositionX(px)} and \textbf{G. ImageMousePositionY(px)} are the x- and y-coordinates of the mouse cursor, respectively, in the WSI image coordinates.
%These columns are empty for the beginning of trial and end of trial rows.
\textbf{H. ViewportUpperLeftX(px)} and \textbf{I. ViewportUpperLeftY(px)} are the x- and y-coordinates of the upper left corner of the \testbedname viewport in the WSI image coordinates.
\textbf{J. ViewportWidth(px)} and \textbf{K. ViewportHeight(px)} are the width and height of the viewport in the WSI image coordinates. 
The \textbf{L. relativeTimestamp(px)} gives the time the fixation occurred relative to the moment the image first appeared on the screen.
\textbf{M. eventType} indicates the type of mouse event that occurred. Its possible values are ``mouse\_move'', ``mouse\_drag'', and ``mouse\_scroll.''
\textbf{N. inMenu}, \textbf{O. inImage}, and \textbf{P. inNavigator} are TRUE or FALSE values indicating whether this fixation fell in the task menu, the displayed image, and the navigator window, respectively. 
%These columns are empty for the beginning of trial and end of trial rows.
\textbf{Q. zoomScale} contains the zoom level of the WSI image, relative to its full size resolution.
\textbf{R. zoomMagnification} stores the zoom level, relative to the fully zoomed-out WSI.

\subsubsection*{Overall Eye-Tracking Metrics Data}
The previous two datasets provided separate gaze and mouse events. This merged perception-action data file integrates both gaze and mouse data for combined analysis, with all events sorted by timestamp. 
Eye-tracking data rows do not contain entries for mouse-specific columns (e.g., MousePositionX(px) and MousePositionY(px)), and vice versa. This file, named \texttt{CombinedGazeMouseData.csv}, is located in the \texttt{\eonedataname} and \texttt{\etwodataname} subdirectories. 

\subsubsection*{Gaze Metrics Data}
The \texttt{GazeMetrics.csv} file in each of the experiment subdirectories include trail-level fixation and saccade measures, such as the total and average duration. The column names are:
\textbf{A.} The \textbf{experimentImageName} is a shortened ID identifying the slide used for the testbed, ranging from ``C1'' to ``C397''. 
\textbf{B.} The {participantID} is a pseudonymous ID identifying the participant, ranging from ``P1'' to ``P10'' for the first protocol and ``P9'' for the second.
\textbf{C.} The \textbf{trialID} is the trial number for this participant. The first image a participant sees will have a \textbf{trialID} of 1, the second image a \textbf{trialID} of 2, and so on, up to 60. 
\textbf{D. totalFixations} is the total number of fixations recorded for this participant for this trial/image viewing. Fewer fixations in the AOI relative to total fixations indicates this search was less efficient.
\textbf{E. totalOrientingFixations} are the fixations before the user began mouse navigation, that is, before the first zoom. It reflects the participant familiarizing themselves with the image before actively searching.
\textbf{F. totalFixationDuration(ms)} and \textbf{G. averageFixationDuration(ms)} are the total and average duration, respectively, of all fixations for this trial in milliseconds. These allow comparison of attention on different stimuli, with longer times suggesting more time analyzing and interpreting the image, or more mental effort.
\textbf{H. totalSaccades} is the total number of saccades recorded for this participant for this trial/image viewing. More saccades indicate more searching and are related to mental workload and cognitive processing. 
\textbf{I. totalSaccadeDuration(ms)} is the total duration of all saccades, also related to mental workload and cognitive processing.
\textbf{J. averageSaccadeDuration(ms)} is the average duration of all saccades.
\textbf{K. averageSaccadeAmplitude(deg)} is the average amplitude of saccades in degrees of visual field, related to physiological arousal and mental workload.

\begin{figure}[!t]
    \centering
    \includegraphics[width=0.98\linewidth]{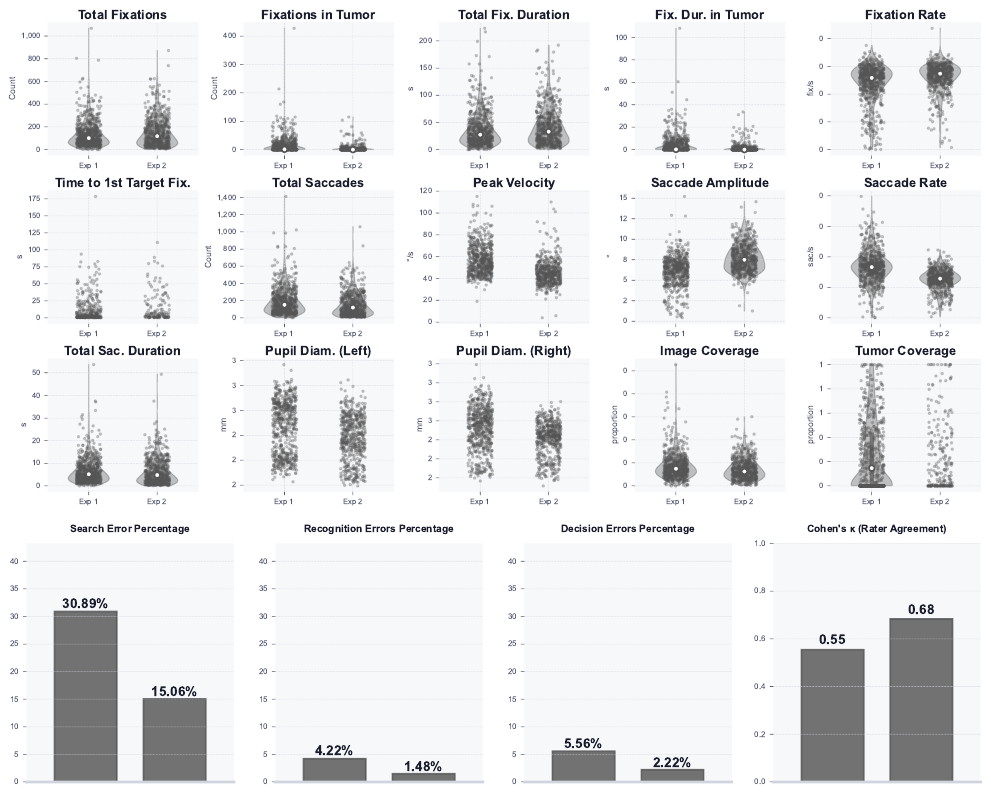}
    \caption{\rv{\textbf{Behavior variables.}
the count of fixations inside or outside of tumors or ROIs, pupil diameter, and the types of errors (search, recognition, and decision). These error types are taken from the radiology literature, with search errors being false negative cases where the viewer never fixated on the tumor,
recognition where the viewer did not fixated on the tumor sufficiently long to recognize it as a ROI (500ms), and decision
where the viewer fixated on the tumor for at least 500ms. These measures can help researchers understand how pathologists navigate and
understand WSIs. The most common false negative source was search errors, with 30.89\% and 15.06\% of cases in the first and second experiment being search errors. Compare this to the second greatest error source, decision errors, at 5.56\% and 2.22\% of cases.}}
    \label{fig:metrics}
\end{figure}

\end{document}